\documentclass[runningheads]{llncs}
\ifdefined\pdfinfoomitdate\pdfinfoomitdate=1 \fi

\usepackage[T1]{fontenc}
\usepackage{graphicx}
\usepackage{amsmath}
\usepackage{amssymb}
\usepackage{array}
\usepackage{booktabs}
\usepackage{multirow}
\usepackage{algorithm}
\usepackage{algpseudocode}
\usepackage{placeins}
\usepackage{xcolor}
\usepackage[figuresright]{rotating}
\usepackage[hidelinks]{hyperref}

\graphicspath{{figures/}}
\newcolumntype{P}[1]{>{\raggedright\arraybackslash}p{#1}}
\newcommand{\sd}[1]{\textcolor{gray}{\,$\pm$\,#1}}
\algrenewcommand\alglinenumber[1]{\scriptsize #1:}

\makeatletter
\renewcommand\floatc@ruled[2]{\small{\@fs@cfont #1} #2\par}
\makeatother

\begin{document}

\title{Speeding up Automated Algorithm Design by Batch Evaluation: A Case Study of Genetic Programming for Ant Colony Optimisation}
\titlerunning{Speeding up AAD by Batch Evaluation}
\author{Bo-Cheng Lin, Yi Mei, Mengjie Zhang}
\authorrunning{Bo-Cheng Lin, Yi Mei, Mengjie Zhang}
\institute{Centre for Data Science and Artificial Intelligence, \\ and the School of Engineering and Computer Science, \\Victoria University of Wellington, Wellington, 6140, New Zealand \\ \{ bocheng.lin, yi.mei, mengjie.zhang \} @ecs.vuw.ac.nz}
\maketitle

\begin{abstract}
Evaluation is becoming the bottleneck of automated algorithm design, especially for iterative algorithms such as ant colony optimisation (ACO), which must run to completion on multiple instances. A transition rule designed by genetic programming reads the construction state of each ant and must be evaluated at every step. However, existing GPU methods for ACO precompute a fixed rule and parallelise the ants of one run. To address this issue, we propose BlockEval, a batch evaluation framework with GPU and CPU backends. The GPU backend executes many runs simultaneously, each as a whole in one thread block, and compiles one module per generation. When an ant has visited all its candidate cities, a warp-cooperative fallback stalls only its warp where possible. On travelling salesman problems with 50--1000 cities, the GPU backend is 4.1--16.5 times faster than the CPU backend on 32 cores and 11.2--17.8 times faster than EvoGP+TensorACO-CL.
\keywords{Automated algorithm design \and Genetic programming \and Ant colony optimisation \and GPU acceleration.}
\end{abstract}

\section{Introduction}
\label{sec:intro}

Automated algorithm design (AAD) uses search instead of human effort to design algorithms. Candidate algorithms can be generated by genetic programming (GP) hyper-heuristics~\cite{burke2013hh,nguyen2017gphh,zhang2021gpbook} or, more recently, by large language models (LLMs)~\cite{romera2024funsearch,ye2024reevo}. Generating a candidate is becoming cheaper, but evaluating it still requires running it to completion on a set of instances. Therefore, the bottleneck of AAD is shifting from designing algorithms to evaluating them~\cite{romera2024funsearch,li2026agentic}. This paper focuses on speeding up each evaluation.

In combinatorial optimisation, evaluation is particularly expensive when the algorithm being designed is itself a population-based iterative algorithm such as ant colony optimisation (ACO). GP has been used to design the transition rules of ACO (GP-ACO)~\cite{runka2009gpaco,lin2025memetic} and its pheromone update~\cite{tavares2010pheromone}. A designed transition rule scores each candidate city of an ant, and it can read the construction state of the ant in addition to the pheromone and the distance of the edge. For example, terminals computed over the remaining candidate cities, such as their average pheromone, significantly improve the designed rules~\cite{lin2025memetic}. Whichever component is designed, the resulting algorithm is evaluated by simulation, which runs a complete ACO on each training instance. A complete run is necessary because the effect of a rule on the pheromone carries over to all later iterations. In addition, GP-evolved scheduling rules generalise better to unseen instances whose sizes are close to those of the training instances~\cite{mei2016reusability,zhu2026generalisation}. To learn rules that work on instances of various sizes, GP-ACO therefore needs more and larger training instances. However, because of the evaluation cost, existing GP-ACO studies can only use small problem sizes and few training instances~\cite{lin2025memetic}, and the primitive set has been reduced to save execution time~\cite{runka2009gpaco}. Other studies reduce the evaluation cost by approximating the fitness with surrogate models or multi-fidelity evaluation~\cite{zhang2021surrogate,zhang2022multifidelity}, or by using a small evaluation budget~\cite{ye2024reevo}. In contrast, this paper speeds up the exact fitness evaluation, and the two can be combined.

A natural way to reduce this cost is to use GPUs, and GPU methods exist for both parts of this evaluation, namely ACO~\cite{cecilia2013gpuaco,yang2024tensoraco} and the program trees of GP~\cite{wu2026evogp}. However, the performance of an algorithm is closely tied to the hardware and software it runs on~\cite{hooker2021lottery}. To make full use of GPU acceleration, the design of an algorithm should take into account both the characteristics of the hardware and the structure of its workload. In GP-ACO, the workload of one generation consists of many independent runs, one for each program on each instance. Each run is a long sequence of dependent construction steps, and in every step the rule designed by GP scores the candidate cities of each ant. As a result, it is not easy to apply existing GPU methods to GP-ACO, even in combination, mainly due to two reasons.

\begin{figure}[!t]
\centering
\includegraphics[width=\textwidth]{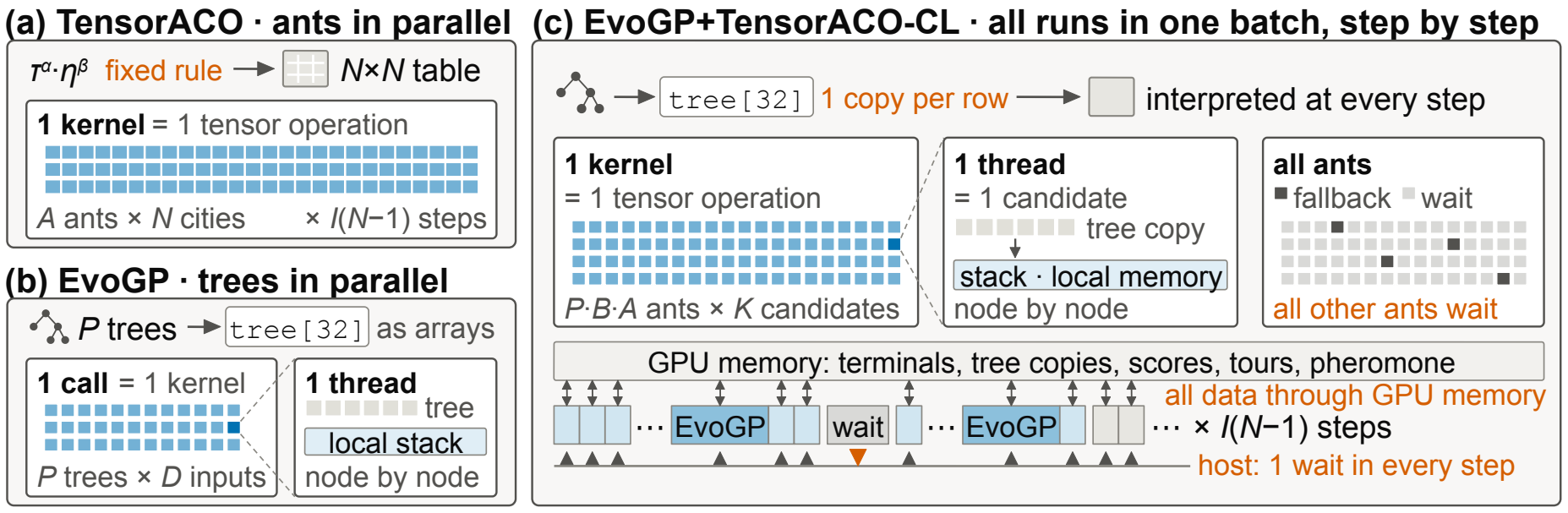}
\vspace{-7mm}
\caption{Existing GPU methods for the two parts of the evaluation, and their combination. Blue cells denote work, dark cells denote ants falling back to the full graph, and light grey cells denote waiting. All three are parallel, but none treats a complete run as the unit of parallelism. (a) TensorACO processes the ants of one run in parallel. Its rule is fixed and is read from a table computed once per iteration, and every ant considers all cities in every step. (b) EvoGP evaluates all trees on a batch of inputs in one call. Each thread copies a tree into local memory and interprets it node by node. (c) EvoGP+TensorACO-CL executes all runs of a generation as one batch, but advances them one construction step at a time, and calls EvoGP at every step with one tree copy per candidate. Double-headed arrows are the reads and writes of the kernels, and the marks on the host line are kernel launches and the points where the host waits. All data pass through GPU memory, the host waits once in every step, and all other ants wait while some fall back. EvoGP+TensorACO executes the same steps without candidate lists, and every ant considers all cities as in (a).}
\label{fig:execution}
\vspace{-7mm}
\end{figure}

First, the transition rule cannot be precomputed. Existing GPU implementations of ACO fix the rule. For example, TensorACO~\cite{yang2024tensoraco} computes the transition weights of all edges once per iteration as a table, and the ants only read this table during construction (Fig.~\ref{fig:execution}(a)). Such a table can only hold a rule that depends on the edge alone. A rule designed by GP also reads the construction state of each ant and needs to be evaluated at every construction step. On a GPU, EvoGP~\cite{wu2026evogp} evaluates all program trees of GP on a batch of inputs in one call (Fig.~\ref{fig:execution}(b)). However, EvoGP runs as a kernel of its own and can only be called from outside the construction loop, once at every construction step (Fig.~\ref{fig:execution}(c)). In each call, the terminals, the scores and a copy of the tree for every candidate need to pass through GPU memory, and the control flow stays on the host.

Second, existing GPU methods for ACO are designed to accelerate one run, whereas a generation consists of many independent runs. GPU implementations of ACO map one run to the whole GPU~\cite{cecilia2018gpuaco}, typically with one thread block or warp per ant~\cite{cecilia2013gpuaco,dawson2013gpuaco,skinderowicz2016gpuacs}, and TensorACO represents the ants of one run as tensors. The runs of a generation can be stacked into one batch, as in GPU frameworks for evolutionary computation such as EvoX~\cite{huang2025evox,liang2026tensor}, MetaDE~\cite{chen2025metade} and AutoPSO~\cite{yu2026autopso}. However, all ants of the batch then advance one construction step at a time. This suits algorithms such as differential evolution (DE) and particle swarm optimisation (PSO), in which a step is one generation of a whole population. In contrast, an ACO run consists of a long sequence of small steps, and every construction step becomes a synchronisation point of all runs (Fig.~\ref{fig:execution}(c)). Without candidate lists, as in TensorACO, every ant reads and scores all cities in every step, and the data moved, the time and the memory of a step all grow with the problem size. With candidate lists, most steps are cheap, but all ants of the batch need to wait when one of them takes a more expensive step.

To the best of our knowledge, no existing study evaluates the designed rules of an iterative algorithm on a GPU with a complete run as the unit of parallelism, and compares the results with existing GPU methods and with an equally optimised multi-core CPU program. Further related work is discussed in Section~\ref{S-sec:related} of the supplementary material.

Both reasons stem mainly from the movement of data. In computer systems, performance is increasingly limited by moving data rather than by arithmetic~\cite{wulf1995wall,ivanov2021data}. GPU acceleration in evolutionary computation is usually obtained by tensorisation, which expresses every step of an algorithm as operations on whole tensors~\cite{liang2026tensor,yang2024tensoraco}. Each operation is a kernel that reads its operands from GPU memory and writes its result back. In a computation with many such kernels, every intermediate result therefore passes through GPU memory~\cite{wahib2014fusion}. Moreover, the threads of a warp execute the same instruction and wait for each other when their control flow diverges~\cite{fung2007warp}. For a long sequence of small dependent steps with expensive branches, these costs can outweigh the parallelism within a step, and a GPU need not be faster than an optimised multi-core CPU~\cite{lee2010debunking}. To reduce the movement of data, we evaluate the programs inside the kernel that constructs the tours of a run, and we treat a complete run as the unit of parallelism, rather than an ant or a construction step (Fig.~\ref{fig:framework}(b)). All construction steps of an iteration are then executed in one kernel, the inputs and outputs of the programs are kept in registers, and many runs are executed simultaneously.

However, this is not straightforward, and there are two main difficulties. First, the GP programs change at every generation. If they are compiled into the kernel, the compilation needs to be repeated at every generation, and compiling GP programs for a GPU is expensive unless the cost is amortised~\cite{ayral2017gpugp}. Second, a kernel that executes a whole run contains divergent control flow, which is known to be costly on a GPU~\cite{laine2013megakernels}. When all cities in the candidate list of the current city have been visited, an ant needs to select the next city from all unvisited cities. We call this step the fallback to the full graph. It is much more expensive than an ordinary step, and when an ant takes this expensive branch, the other ants need to wait. With many synchronised ants, at least one of them takes this branch in a large fraction of the construction steps.

The goal of this paper is to design a batch evaluation framework for GP-ACO, named BlockEval, whose GPU backend reduces the evaluation time per generation without changing the fitness computation. Specifically, this paper has the following three objectives.
\begin{enumerate}
\vspace{-1.5mm}
\item To analyse why existing GPU methods are not suitable for this workload, and to verify the analysis by experiments on TensorACO and on two baselines that combine it with EvoGP, without and with candidate lists.
\item To design a batch evaluation framework that treats one run of one program on one instance as the unit of parallelism and reduces the movement of data between the host, GPU memory and the threads. Its GPU backend should execute each run as a whole, compile all programs of a generation into one module, and make full use of the GPU by executing many runs simultaneously. We also measure its speedup over the CPU backend, the effect of each design choice, its scalability and the cost of compiling at every generation.
\item To design a fallback to the full graph such that, when an ant takes this expensive branch, not all other ants need to wait, and to measure its effect at different problem sizes.
\end{enumerate}

\vspace{-2mm}

Although this paper uses ACO for the travelling salesman problem (TSP) as a case study, the analysis and the corresponding techniques also apply to other iterative algorithms that call a designed rule in a long sequence of dependent steps with expensive branches, such as constructive heuristics for routing or scheduling.

\vspace{-4mm}
\section{Background}
\label{sec:background}

\vspace{-2mm}
\subsection{ACO for TSP}
\label{sec:background:tsp}

Given $N$ cities and the distance $d_{ij}$ between each pair of them, the TSP~\cite{applegate2006tsp} aims to find the shortest tour that visits each city exactly once and returns to the starting city.

ACO~\cite{dorigo2004book} uses a colony of ants to construct tours repeatedly and records the search experience in the pheromone $\tau_{ij}$. In each iteration, each of the $A$ ants starts from a city, and an ant at city $i$ selects the next city $j$ with probability
\vspace{-2mm}
\begin{equation}
\Pr(j\mid i)=\frac{\tau_{ij}^{\alpha}\eta_{ij}^{\beta}}{\sum_{k\in C}\tau_{ik}^{\alpha}\eta_{ik}^{\beta}},\qquad j\in C,
\label{eq:aco}
\vspace{-2mm}
\end{equation}
where $\eta_{ij}=1/d_{ij}$ is the heuristic information, and $\alpha$ and $\beta$ control the relative weights of the pheromone and the heuristic information. To reduce the computational cost, a candidate list of the $K$ nearest neighbours is kept for each city, and the candidate set $C$ contains only the unvisited cities in this list. If all cities in the candidate list have been visited, the ant selects the unvisited city with the largest transition weight~\cite{dorigo2004book}, which requires scanning all $N$ cities. This step is the fallback to the full graph. This paper considers three ACO variants, namely Ant System (AS)~\cite{dorigo1996as}, Ant Colony System (ACS)~\cite{dorigo1997acs} and MAX--MIN Ant System (MMAS)~\cite{stutzle2000mmas}. They differ mainly in how the pheromone is updated.

\vspace{-2mm}
\subsection{GP to Learn ACO Transition Rules}
\label{sec:background:gpaco}

GP-ACO uses GP to design the transition rules of ACO automatically~\cite{runka2009gpaco,lin2025memetic}. Each GP individual is a program tree that takes the features of a candidate city as inputs and outputs a score for the city. The ant then selects the next city based on these scores. The terminal set includes features of the candidate edge (e.g., its distance and pheromone), the construction state of the ant and the progress of the run (Table~\ref{S-tab:language} in the supplementary material). The fitness of a program $p$ is the average gap of the resulting ACO over $B$ training instances,
\vspace{-2mm}
\begin{equation}
f(p)=\frac{1}{B}\sum_{b=1}^{B}100\,\frac{L_{pb}-L_b^{*}}{L_b^{*}},
\label{eq:fitness}
\vspace{-2mm}
\end{equation}
where $L_{pb}$ is the length of the best tour found on instance $b$ and $L_b^{*}$ is its reference optimal length. In each generation, the fitness needs to be computed for all $P$ programs, which requires $P\times B$ runs of ACO.

A rule that depends only on the edge can be precomputed as a table, as is done for the heuristic matrices of DeepACO~\cite{ye2023deepaco} and the transition weights of TensorACO. However, seven of the ten terminals depend on the construction state of the ant, such as DistRank, the distance rank of a candidate among the unvisited cities. We trained GP 10 times each on TSP100 and TSP500, and all 20 best rules read at least one of these terminals (Table~\ref{S-tab:language} in the supplementary material).

\vspace{-3mm}

\section{The Proposed Framework: BlockEval}
\label{sec:framework}

\vspace{-2mm}
\subsection{Overall Framework}
\label{sec:framework:structure}

\begin{figure}[!t]
\centering
\includegraphics[width=\textwidth]{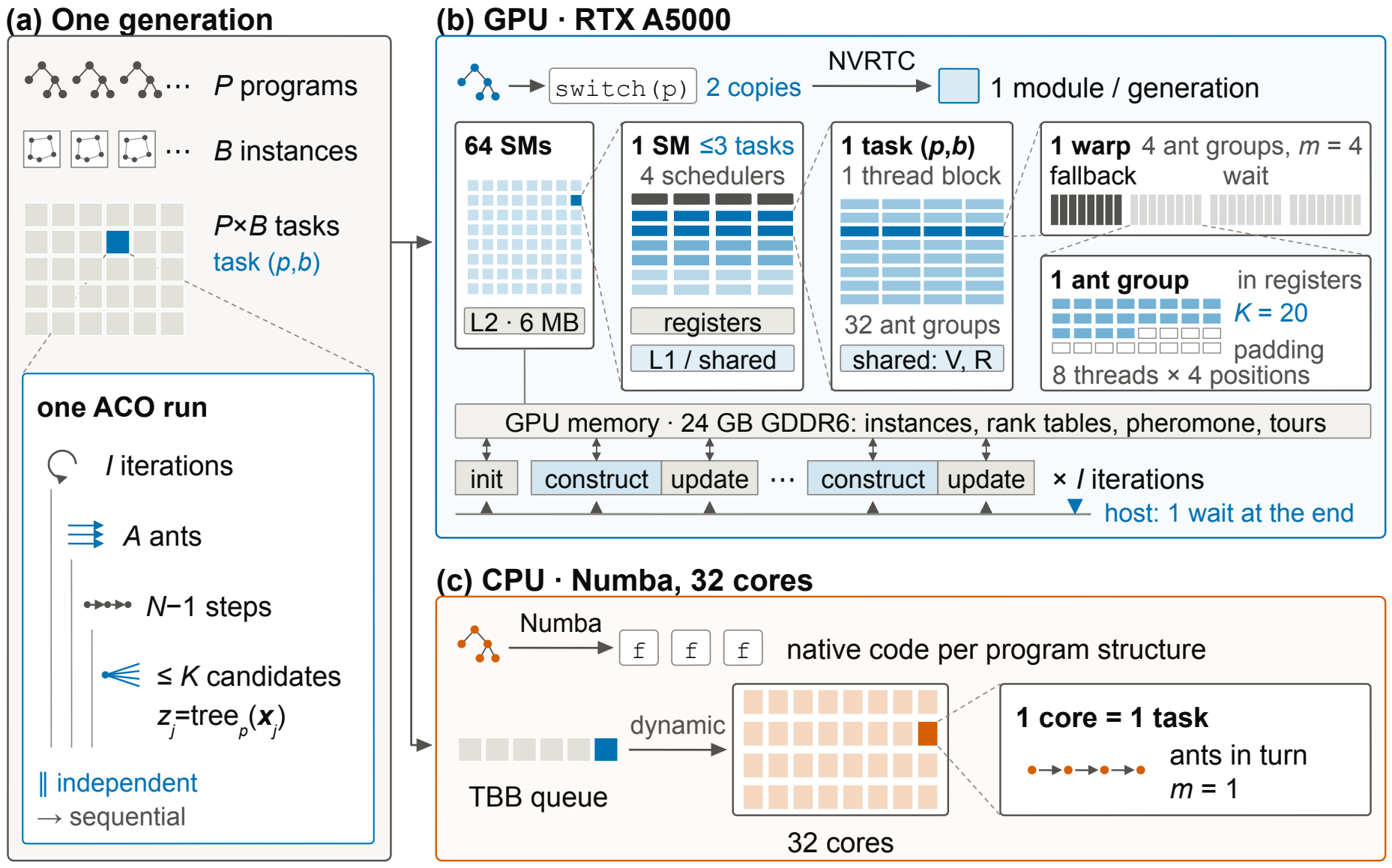}
\vspace{-7mm}
\caption{Overall framework of BlockEval. (a) One generation consists of $P\times B$ tasks, each of which runs ACO once. (b) In the GPU backend, all programs of a generation are compiled into one module by NVRTC, the runtime compilation library of CUDA. Each task is executed by a thread block on a streaming multiprocessor (SM). The ant groups in a warp are synchronised, and when one group falls back to the full graph, the other groups need to wait. The arrows between the kernels and the GPU memory are the reads and writes of a kernel, and the marks on the host line are the kernel launches and the single point where the host waits. (c) In the CPU backend, the Threading Building Blocks (TBB) library assigns the tasks dynamically to 32 cores.}
\label{fig:framework}
\vspace{-6mm}
\end{figure}

The evaluation of one generation consists of $P\times B$ tasks. A task $(p,b)$ is a complete run of the ACO controlled by program $p$ on instance $b$. The tasks are independent of each other (Fig.~\ref{fig:framework}(a)). Within a task, the iterations depend on each other through the pheromone, and the construction steps depend on each other through the visited cities. Only the ants and the candidate cities in the same step can be processed in parallel, and the GP program is evaluated at the level of candidate cities. Each task therefore has about $A\times K$ parallel operations per step, whereas the $I(N-1)$ selections of each ant need to be made one after another, where $I$ is the number of ACO iterations.

Based on this structure, BlockEval treats a task as the basic unit of parallelism and executes each run as a whole rather than step by step. As shown in Fig.~\ref{fig:framework}(b) and (c), it has two backends that follow the same computation rules (Section~\ref{sec:setup}). In the GPU backend, each task is executed by one thread block, and the ants and candidate cities of the task are assigned to the threads of this block (Sections~\ref{sec:framework:gpu} and \ref{sec:framework:fallback}). Compared with EvoGP+TensorACO-CL (Fig.~\ref{fig:execution}(c)), this mapping reduces the movement of data. The instances and the compiled programs are sent to the GPU once per generation, and the host waits only once, after all iterations. In an ordinary step, an ant reads only the entries of its $K$ candidates, which are stored contiguously for each city. The programs are compiled into the kernel, and the terminals and the outputs of the programs stay in registers instead of passing through GPU memory. In the CPU backend, each task is executed by one core (Section~\ref{sec:framework:cpu}).

\vspace{-4mm}
\subsection{Residual Form of the Transition Rule}
\label{sec:framework:residual}
The transition rule takes a residual form, which is part of the GP-ACO training problem and is the same for all implementations compared in this paper. For each candidate city $j$, the program tree computes $z_j=\mathrm{tree}(\boldsymbol{x}_j)$, where $\boldsymbol{x}_j$ denotes the values of the terminals at candidate $j$. Instead of replacing the original transition weight with $z_j$, we use $z_j$ to modify it by a bounded multiplier,
\vspace{-2mm}
\begin{equation}
w_j=b_j\,\bigl[1+\gamma\tanh(z_j)\bigr],\qquad b_j=\tau_{ij}\,\eta_{ij}^{2},\quad \gamma=\tfrac13 .
\label{eq:transition}
\vspace{-2mm}
\end{equation}
Here $b_j$ is the original weight with $\alpha=1$ and $\beta=2$, and $w_j$ replaces the numerator of Eq.~(\ref{eq:aco}). A tree that always outputs 0 gives exactly the original ACO, and GP therefore modifies the original rule rather than replacing it. In addition, the evaluation cost does not decrease when GP finds a trivial tree, and the measured speedup comes from the evaluation method, not from the rule.

\vspace{-2mm}
\subsection{GPU Kernel}
\label{sec:framework:gpu}

A thread block (Fig.~\ref{fig:framework}(b)) contains $\min(A,32)$ ant groups, each with $T$ threads, so that each thread needs to read only part of the $K$ candidates of the current city. When $A>32$, the remaining ants are processed in turn. In each iteration, a construction kernel is launched, followed by an update kernel, and the iterations are synchronised at the boundary between the two kernels. During construction, each thread reads its candidates at fixed positions and masks the visited cities, which avoids branching. The full pseudo-code of the construction kernel is given in Algorithm~\ref{S-alg:construct} in the supplementary material.

\vspace{-3mm}

\paragraph{Compiling Once per Generation.} At the beginning of each generation, each program is translated into straight-line code and placed into a \texttt{switch} statement indexed by the program number. To amortise the compilation cost~\cite{ayral2017gpugp}, all programs of the generation are compiled into one module by NVRTC~\cite{nvidia2026nvrtc}, the runtime compilation library of CUDA. Each thread scores its candidate positions in a loop, which we call the scoring loop. The candidate list is padded to 32 positions, and each thread has $32/T$ positions. Unrolling this loop would copy the program for each of these positions. With one more copy in the fallback, each program would appear $32/T+1$ times in the kernel. We therefore keep the loop, and each program appears only twice, once in the selection from the candidate list (the ordinary selection) and once in the fallback to the full graph.

\vspace{-2mm}

\subsection{Warp-Cooperative Fallback}
\label{sec:framework:fallback}

The threads of a GPU warp execute the same instruction simultaneously. When one ant group enters the fallback to the full graph (Section~\ref{sec:background:tsp}), the other synchronised groups therefore need to wait. Let $q$ be the probability that an ant falls back to the full graph in a construction step, and let $m$ be the number of synchronised ant groups. If the groups fall back independently, the fraction of waiting steps, in which the synchronised groups need to wait, is
\vspace{-2mm}
\begin{equation}
\pi_m(q)=1-(1-q)^{m} .
\label{eq:amplify}
\vspace{-2mm}
\end{equation}

To address this issue, we propose the warp-cooperative fallback (WCF), which consists of two components (Fig.~\ref{fig:framework}(b)). First, the $T$ threads of an ant group enumerate the unvisited cities cooperatively. Thread $\ell$ handles the cities whose indices are congruent to $\ell$ modulo $T$, and the results of the threads are merged at the end. DistRank is obtained by counting bits in a bitmap (lines 6 and 9 of Algorithm~\ref{alg:wcf}). The rank $r_i(j)$ of each city $j$ by its distance from city $i$ is precomputed once per instance. The DistRank of $j$ is then the number of unvisited cities ranked before $j$. In this way, the cost of the fallback is proportional to the number of unvisited cities rather than to $N$. Second, for AS and MMAS, the construction steps are not synchronised across the whole thread block. Only the ant groups in the same warp of 32 threads are synchronised, i.e., $m=32/T$ rather than 32. The local update of ACS requires the whole thread block to be synchronised in every step, and $m=32$ for ACS. Algorithm~\ref{alg:wcf} shows one construction step of an ant group with WCF.

\vspace{-6mm}
\begin{algorithm}[H]
\caption{One construction step with WCF}
\label{alg:wcf}
\scriptsize
\begin{algorithmic}[1]
\Require thread $\ell\in\{0,\dots,T-1\}$ of an ant group, current city $i$, visited bitmap $V$ in shared memory
\Ensure next city $j^{*}$
\If{the candidate list of $i$ contains an unvisited city}
  \State Select $j^{*}$ from the candidate list \Comment{ordinary selection}
\Else \Comment{fallback to the full graph}
  \State $U_\ell\gets\{j : V[j]=0,\ j\equiv\ell\pmod T\}$ \Comment{read from $V$}
  \State Compute the mean and variance of each terminal over all unvisited cities (sums over $U_\ell$, then over the $T$ threads)
  \State $R\gets 0$; set bit $r_i(j)$ of $R$ for each $j\in U_\ell$; synchronise the $T$ threads
  \State $(w_\ell,j_\ell)\gets(-\infty,\mathrm{none})$
  \For{$j\in U_\ell$ in increasing order}
    \State $\mathrm{DistRank}(j)\gets$ number of set bits of $R$ before bit $r_i(j)$
    \State Compute $z_j$ and $w_j$ by Eq.~(\ref{eq:transition}); \textbf{if} $w_j>w_\ell$ \textbf{then} $(w_\ell,j_\ell)\gets(w_j,j)$
  \EndFor
  \State $j^{*}\gets$ the $j_\ell$ with the largest $w_\ell$ over the $T$ threads \Comment{ties go to the smaller $j$}
\EndIf
\State Thread 0 appends $j^{*}$ to the tour and sets $V[j^{*}]\gets 1$
\State Synchronise only the $T$ threads (AS, MMAS) or the whole block (ACS, then the local update)
\end{algorithmic}
\end{algorithm}
\vspace{-2mm}

\vspace{-8mm}
\subsection{CPU Backend}
\label{sec:framework:cpu}

The CPU backend evaluates the same tasks with Numba~\cite{lam2015numba} (Fig.~\ref{fig:framework}(c)). The whole ACO loop is compiled into native code once. The CPU backend also compiles at every generation, one scoring function for each program structure. In the fallback, it scans all cities once to collect the unvisited ones, and obtains DistRank from the precomputed ranks, or by pairwise comparisons when few cities are unvisited. The tasks are assigned dynamically to the cores by the Threading Building Blocks (TBB) library~\cite{voss2019tbb}.

\vspace{-4mm}

\section{Experiment Design}
\label{sec:setup}

\vspace{-2mm}

The experiments aim to answer three research questions, which correspond to the three objectives. RQ1: How much time and GPU memory do existing GPU methods need on this workload, with and without candidate lists, and is this consistent with the analysis in Section~\ref{sec:intro}? RQ2: How much faster is the GPU backend of BlockEval than the CPU backend on 32 cores, and how much does each of its design choices contribute? How does the speedup change with the problem size and the numbers of programs, instances and ants, and how much does compiling at every generation cost? RQ3: How much does WCF reduce the evaluation time, and how does its effect change with the problem size?

\vspace{-2mm}

\paragraph{Datasets.} The experiments use Euclidean TSP instances from ML4CO-Bench-101~\cite{ma2025ml4co}, with 50, 100, 200, 500 and 1000 cities uniformly distributed in the unit square. The optimal tour lengths of these instances are used as the reference lengths. Each ACO run has $I=500$ iterations and a candidate list of $K=20$ cities, without local search. All parameter settings are listed in Table~\ref{S-tab:setup}.

\vspace{-3mm}

\paragraph{Time Measurement.} To compare the methods on the same workload, each of the 10 runs uses the recorded populations of 10 consecutive generations of a separate GP run (256 programs, TSP100). These populations are evaluated under every configuration, with the instances and random-number keys of that configuration. The default configuration at each problem size uses $P=B=A=32$ and $T=8$. The two backends follow the same computation rules, with the same counter-based random numbers and deterministic single-precision arithmetic, and they differ only in rounding. The average gap of the same program differs between the two backends by 0.005 percentage points on average and by at most 0.23 (Table~\ref{S-tab:consistency}). We therefore compare only the evaluation time.

\vspace{-3mm}

\paragraph{Compared Methods.} Hardware acceleration should be measured against an optimised CPU program~\cite{chitty2012gp,crary2025fpga}. We therefore compare the GPU backend with the CPU backend, and also with existing GPU methods. On one core, the CPU backend is 95 times faster than the sequential Python implementation used in existing GP-ACO studies~\cite{lin2025memetic} (Fig.~\ref{S-fig:cpu_baseline}). TensorACO~\cite{yang2024tensoraco} is run with its own fixed rule and without programs, as a reference. Its released code processes one run at a time. We stack the runs of a generation into one batch, which makes it 28--88 times faster (Table~\ref{S-tab:nocl}). Two baselines, EvoGP+TensorACO and EvoGP+TensorACO-CL, evaluate the designed rules by calling EvoGP~\cite{wu2026evogp}, the state-of-the-art GPU evaluator of GP program trees, at every construction step. EvoGP+TensorACO multiplies the precomputed weights of TensorACO by the outputs of the programs as in Eq.~(\ref{eq:transition}). EvoGP+TensorACO-CL (Fig.~\ref{fig:execution}(c)) adds candidate lists and the fallback to the full graph. It runs AS with the same candidate lists, selection probabilities and random numbers as BlockEval and evaluates the same workload. We report the fastest of its three execution modes (Table~\ref{S-tab:compose}). The times of the existing GPU methods exclude their one-time compilation.

\vspace{-3mm}

\paragraph{Platform and Statistics.} Unless otherwise stated, the GPU experiments are conducted on an NVIDIA RTX A5000, and the CPU experiments on a node of an AMD EPYC 9634 cluster with 32 physical cores used exclusively. The evaluation time per generation is the wall-clock time from when the input of a generation is ready to when the fitness values are obtained, including compilation, data transfer and search. The cost per transition (ns) is the evaluation time divided by the $PBAI(N-1)$ city selections in a generation. Each configuration is run 10 times independently, and we report the mean and the standard deviation, with speedups computed from paired runs.

\vspace{-4mm}

\section{Results and Discussions}
\label{sec:results}

\vspace{-2mm}

\subsection{Evaluation Time and Memory}
\label{sec:results:evogp}

\vspace{-2mm}

Table~\ref{tab:evogp} shows the evaluation time per generation and the peak GPU memory under the default configuration of AS. The GPU backend is the fastest at all five problem sizes and uses the least GPU memory. For RQ1, the GPU backend is 14.6--156$\times$ faster than EvoGP+TensorACO and 11.2--17.8$\times$ faster than EvoGP+TensorACO-CL. Neither of these two baselines is faster than the CPU backend on 32 cores. Even TensorACO with its own fixed rule, which only reads a precomputed table, is not faster than the GPU backend. At $N=50$, the two take a similar time, but at $N=1000$ the GPU backend is 9.9$\times$ faster. For RQ2, the GPU backend is 7.4--14.8$\times$ faster than the CPU backend on 32 cores for AS. The time of the GPU backend includes 1.6--1.8\,s of compilation per generation, which is more than half of the evaluation time at $N=50$ but only 1.4\% at $N=1000$. Fig.~\ref{fig:scaling}(a) shows the same pattern for ACS and MMAS, for which the GPU backend is 4.1--11.8$\times$ and 8.2--16.5$\times$ faster, respectively. Tables~\ref{S-tab:ratios} and~\ref{S-tab:cross_gpu} give the paired ratios and the results on six other GPUs.

\begin{table}[!h]
\vspace{-6mm}
\centering
\caption{Evaluation time per generation (s) and peak GPU memory (GB) under the default configuration of AS ($P=B=A=32$). TensorACO runs its own fixed rule without programs as a reference. The times are the mean \sd{standard deviation} over 10 runs and exclude the one-time compilation of the existing GPU methods. Comp.\ is the compilation time per generation of the GPU backend (s), which is included in its time. The memory is the sampled peak held by the process, and for a generation executed in batches ($\dagger$) it is that of one batch. The best value in each row is in bold.}
\label{tab:evogp}
\vspace{-2mm}
\fontsize{6.5}{8}\selectfont
\begingroup\setlength{\tabcolsep}{1.0pt}
\begin{tabular}{ccccccccccc}
\toprule
\multirow{3}{*}{$N$} & \multicolumn{6}{c}{Existing GPU methods} & \multicolumn{4}{c}{\textbf{BlockEval (Ours)}} \\
\cmidrule(lr){2-7}\cmidrule(lr){8-11}
 & \multicolumn{2}{c}{\shortstack{TensorACO\\(fixed rule)}} & \multicolumn{2}{c}{\shortstack{EvoGP+\\TensorACO}} & \multicolumn{2}{c}{\shortstack{EvoGP+\\TensorACO-CL}} & \shortstack{CPU\\(32 cores)} & \multicolumn{3}{c}{GPU} \\
\cmidrule(lr){2-3}\cmidrule(lr){4-5}\cmidrule(lr){6-7}\cmidrule(lr){8-8}\cmidrule(lr){9-11}
 & Time & Mem. & Time & Mem. & Time & Mem. & Time & Time & Comp. & Mem. \\
\midrule
50 & 3.21\sd{0.05} & 1.73 & 44.83\sd{1.94} & 3.04 & 44.33\sd{1.67} & 1.66 & 43.34\sd{2.74} & \textbf{3.07}\sd{0.17} & 1.77 & \textbf{0.69} \\
100 & 9.67\sd{0.17} & 4.91 & 172.5\sd{8.6} & 7.18 & 92.77\sd{5.37} & 1.78 & 77.23\sd{8.01} & \textbf{5.23}\sd{0.23} & 1.82 & \textbf{0.76} \\
200 & 33.84\sd{0.41} & 17.66 & 694.3\sd{34.0} & 20.63 & 192.4\sd{11.7} & 2.15 & 129.0\sd{8.6} & \textbf{11.27}\sd{0.41} & 1.61 & \textbf{1.03} \\
500 & 250.2\sd{1.8} & 17.66$^{\dagger}$ & 4477.6\sd{198.3} & 20.04$^{\dagger}$ & 514.6\sd{18.9} & 6.35 & 351.0\sd{34.1} & \textbf{39.57}\sd{2.38} & 1.64 & \textbf{2.79} \\
1000 & 1117.7\sd{2.0} & 17.66$^{\dagger}$ & 17,691.7\sd{856.9} & 11.64$^{\dagger}$ & 1265.5\sd{25.9} & 19.79 & 839.0\sd{76.1} & \textbf{113.8}\sd{7.1} & 1.58 & \textbf{8.95} \\
\bottomrule
\end{tabular}
\endgroup
\vspace{-8mm}
\end{table}

\vspace{-4mm}
\subsection{Why Existing GPU Methods Are Slower}
\label{sec:results:why}

\vspace{-2mm}

\paragraph{Without Candidate Lists.} To explain these results (RQ1), we measure the time of one construction step, in which each of the 32,768 ants of a generation selects one city (Table~\ref{tab:steps}). For TensorACO, which has no candidate lists, the time per step grows in proportion to $N$, from 0.13\,ms to 2.2\,ms. In every step, each of its ants reads the weights and the visited flags of all $N$ cities, $8N$ bytes in total, which is $N/K=2.5$--50 times as much data as with candidate lists. In contrast, a step of the GPU backend takes 0.11--0.23\,ms and a step of the CPU backend 1.3--1.8\,ms from $N=50$ to 1000. TensorACO is therefore faster than the CPU backend up to $N=500$ but slower at $N=1000$, although it evaluates no program (Table~\ref{tab:evogp}). EvoGP+TensorACO moves even more data, since the terminals, the scores and a copy of the program pass through GPU memory for every city in every step. A generation of EvoGP+TensorACO takes 14.0--20.5 times longer than one of TensorACO with its fixed rule (Table~\ref{tab:evogp}), and most of this time does not depend on what the programs compute. At $N=1000$, a batch whose program is a single terminal still needs 23.5\,ms per step (Table~\ref{S-tab:nodes}), against 35.4\,ms on average.

\begin{table}[!t]
\centering
\caption{One construction step under the default configuration of AS. The time per step is the time per generation divided by the $I(N-1)$ construction steps of a run, as means over 10 runs. The shares are measured in one generation, as the time for which the kernels run on the GPU divided by the time of a step (Table~\ref{S-tab:stages}). The last row is the percentage of steps in which at least one ant of EvoGP+TensorACO-CL falls back.}
\label{tab:steps}
\vspace{-2mm}
\scriptsize
\begingroup\setlength{\tabcolsep}{3pt}
\begin{tabular}{lccccc}
\toprule
 & $N=50$ & 100 & 200 & 500 & 1000 \\
\midrule
\multicolumn{6}{l}{Time per construction step (ms)} \\
\quad TensorACO (fixed rule) & 0.131 & 0.195 & 0.340 & 1.00 & 2.24 \\
\quad EvoGP+TensorACO & 1.83 & 3.49 & 6.98 & 17.9 & 35.4 \\
\quad EvoGP+TensorACO-CL & 1.81 & 1.87 & 1.93 & 2.06 & 2.53 \\
\quad \textbf{BlockEval (Ours)}, CPU backend (32 cores) & 1.77 & 1.56 & 1.30 & 1.41 & 1.68 \\
\quad \textbf{BlockEval (Ours)}, GPU backend & 0.125 & 0.106 & 0.113 & 0.159 & 0.228 \\
\midrule
\multicolumn{6}{l}{Share of the time of a step of EvoGP+TensorACO-CL (\%)} \\
\quad GPU idle, waiting for the host to launch the next kernel & 46.4 & 46.1 & 36.4 & 22.3 & 10.1 \\
\quad Computing the terminals & 10.4 & 10.3 & 11.2 & 10.7 & 8.9 \\
\quad Programs evaluated by EvoGP & 27.0 & 25.7 & 26.2 & 23.8 & 19.0 \\
\quad Fallback to the full graph & 5.5 & 8.7 & 15.8 & 31.8 & 51.6 \\
\quad Other kernels & 10.8 & 9.2 & 10.4 & 11.4 & 10.4 \\
\midrule
EvoGP+TensorACO-CL, steps with a fallback (\%) & 59.2 & 79.8 & 89.9 & 95.9 & 97.9 \\
\bottomrule
\end{tabular}
\endgroup
\vspace{-4mm}
\end{table}

\vspace{-3mm}

\paragraph{With Candidate Lists.} EvoGP+TensorACO-CL scores the same 655,360 rows in every step regardless of $N$, and a step takes 1.8--2.5\,ms. On the same GPU, the GPU backend makes the same selections in less than a tenth of this time, including its compilation. As shown in the lower part of Table~\ref{tab:steps}, the GPU is idle for 36--46\% of a step of EvoGP+\allowbreak TensorACO-CL for $N\le 200$, waiting for the host to launch the next kernel. This is because the host can launch the kernels of the fallback only after the GPU has reported how many ants have exhausted their candidate lists. In addition, the call to EvoGP takes 19--27\% of a step, and computing its inputs takes 9--11\%. The call alone takes longer than a complete step of the GPU backend. Moreover, all 32,768 ants advance in lockstep, and 59--98\% of the steps contain a fallback, during which all other ants wait. This fallback takes 5.5\% of a step at $N=50$ and 51.6\% at $N=1000$. In the GPU backend, the host waits only once per generation, and a fallback stalls only one warp.

\vspace{-3mm}

\paragraph{Comparison with the CPU Backend.} Although EvoGP+TensorACO-CL runs on a GPU, the CPU backend on 32 cores is at least as fast at all five problem sizes (Table~\ref{tab:steps}). Each core of the CPU backend executes complete runs by itself with the rule compiled into its loop. The terminals and scores stay in registers, and no data are passed between separate tools or between the host and a GPU. A fallback delays only the ant that takes it. This indicates that executing complete runs is more effective for this workload than executing the steps of all runs on a GPU. The GPU backend applies the same principle with one run per thread block.

\vspace{-2mm}

\subsection{Ablation Study}
\label{sec:results:ablation}

\vspace{-2mm}

We change one design choice of the GPU backend at a time on TSP500, under the default configuration of AS, and Table~\ref{tab:ablation} shows the results. For the design choices of objective 2 (RQ2), executing one ant per thread block instead of one run takes 1.70 times as long. The interpreter kernel avoids compilation but takes 1.38 times as long. It still evaluates the rule inside the construction kernel. EvoGP+TensorACO-CL evaluates the same rule outside this kernel, and the GPU backend is 13.0$\times$ faster than it at this problem size (Table~\ref{tab:evogp}). This indicates that the gain comes mainly from where the rule is evaluated rather than from compiling it. Unrolling the scoring loop (Section~\ref{sec:framework:gpu}) makes each program appear five times in the kernel instead of twice, which slows down the compilation and can reduce the number of resident tasks per SM from three to two, and this variant takes 1.15 times as long. When the runs are executed one at a time, as in existing GPU implementations of ACO~\cite{delevacq2013gpuaco,skinderowicz2020gpummas}, the GPU backend is 40$\times$ faster, because each kernel launch then uses only a small part of the GPU.

For the fallback to the full graph (RQ3), the GPU backend is 8.30$\times$ faster than the direct port, which implements the sequential program on the GPU without WCF (Section~\ref{sec:analysis:wcf}). With four threads per ant group, each thread reads more candidates and more ant groups are synchronised. This variant takes 1.60 times as long.

\begin{table}[!h]
\vspace{-6mm}
\centering
\caption{Ablation study on TSP500 (AS, $P=B=A=32$, RTX A5000). Each variant changes one design choice of the GPU backend, and the time ratio is its evaluation time per generation divided by that of the GPU backend. All variants are listed in Table~\ref{S-tab:ablation}.}
\vspace{-2mm}
\label{tab:ablation}
\scriptsize
\begingroup\setlength{\tabcolsep}{8pt}
\begin{tabular}{llc}
\toprule
Design choice of the GPU backend & Variant & Time ratio \\
\midrule
One run per thread block & One ant per thread block & 1.70 \\
Programs compiled into the kernel & Interpreter & 1.38 \\
Scoring loop kept & Unrolled scoring loop & 1.15 \\
All runs executed simultaneously & One run at a time & 40.0 \\
\midrule
WCF & Direct port & 8.30 \\
8 threads per ant group & 4 threads per ant group & 1.60 \\
\bottomrule
\end{tabular}
\endgroup
\vspace{-8mm}
\end{table}

\vspace{-4mm}

\section{Further Analysis}
\label{sec:analysis}

\vspace{-2mm}

\subsection{Scalability in Each Dimension}
\label{sec:analysis:scaling}
\vspace{-2mm}

\begin{figure}[!tb]
\centering
\includegraphics[width=\textwidth]{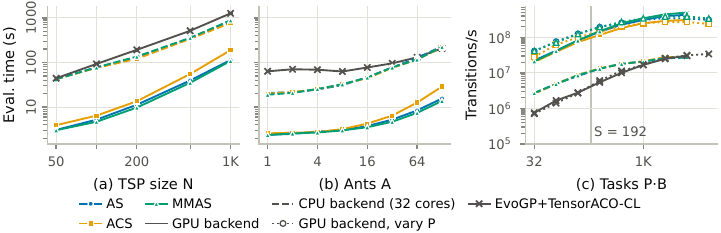}
\vspace{-6mm}
\caption{Evaluation time per generation against (a) the problem size $N$ and (b) the number of ants $A$. (c) Throughput against the number of tasks per generation $PB$ ($N=100$, $A=32$). Solid lines correspond to varying $B$ and dotted lines to varying $P$ (only $B$ for the CPU backend). The vertical line is the number of resident tasks $S=192$. Colours denote the ACO variants. Dashed lines show the CPU backend on 32 cores, and the other coloured lines the GPU backend. The grey line with crosses shows EvoGP+TensorACO-CL for AS, excluding its one-time compilation, with medians over three runs in (b) and (c). Error bars show the standard deviation over 10 runs.}
\label{fig:scaling}
\vspace{-4mm}
\end{figure}

For RQ2, we change one dimension at a time, starting from $N=100$ and $P=B=A=32$ (Fig.~\ref{fig:scaling}). The three ACO variants show the same pattern. As $N$ increases, the evaluation time of both backends grows roughly in proportion to the workload (Fig.~\ref{fig:scaling}(a)). When the number of programs $P$, instances $B$ or ants $A$ increases, the CPU time grows roughly linearly. In contrast, the GPU time stays almost constant for small values and then grows gradually, because the compilation time does not depend on the number of instances or ants, and the GPU is not yet fully used. As shown in Fig.~\ref{fig:scaling}(c), the throughput of the GPU increases with the number of tasks per generation $PB$ and then levels off. It reaches 90\% of its maximum at about 1,024 tasks. This is about five times the $S=192$ tasks that can reside on the A5000 simultaneously. The 32 cores of the CPU backend are fully used with tens of tasks per generation, whereas the A5000 needs about a thousand. Over all 111 configurations, the GPU backend is 1.70--18.0$\times$ faster than the CPU backend. At all points of the grey line, it is 8.8--55.6$\times$ faster than EvoGP+TensorACO-CL (Table~\ref{S-tab:ratios}).

\vspace{-3mm}
\subsection{Effect of WCF at Different Problem Sizes}
\label{sec:analysis:wcf}

\begin{figure}[!tb]
\centering
\begin{minipage}[t]{0.45\textwidth}
\centering
\includegraphics[width=\linewidth]{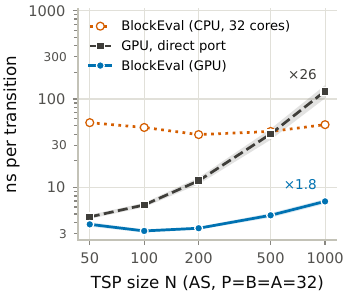}\\[2pt]
\vspace{-1mm}
{\small (a) Cost per transition}
\end{minipage}\hfill
\begin{minipage}[t]{0.45\textwidth}
\centering
\includegraphics[width=\linewidth]{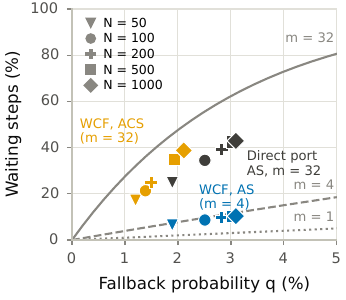}\\[2pt]
\vspace{-1mm}
{\small (b) Fraction of waiting steps}
\end{minipage}
\vspace{-2mm}
\caption{Effect of the fallback to the full graph (RTX A5000, $P=B=A=32$). (a) Cost per transition of AS. The shading shows the standard deviation, and the numbers give the increase from $N=50$ to 1000. (b) Fraction of waiting steps. The grey lines show Eq.~(\ref{eq:amplify}) for $m$ synchronised ant groups. The markers show the measured fraction of steps in which a warp ($m=4$) or a thread block ($m=32$) contains a fallback, over three generations of one run, against the measured $q$. Colours denote the method and the ACO variant, and shapes denote $N$.}
\label{fig:kernel_ablation}
\vspace{-6mm}
\end{figure}

To investigate how much WCF helps (RQ3), we compare the GPU backend with the direct port (Section~\ref{sec:results:ablation}), which also executes each run in one thread block. Fig.~\ref{fig:kernel_ablation}(a) shows that, from $N=50$ to 1000, the cost per transition of the direct port grows by a factor of 26, whereas that of the GPU backend grows by a factor of only 1.8. Compared with the CPU backend, the direct port is faster up to $N=500$ but slower at $N=1000$. This shows that executing each run as a whole is not sufficient.

This difference is mainly due to the fallback to the full graph. For AS, an ant falls back in $q=1.9$--$3.1\%$ of its construction steps. Since the direct port synchronises the whole thread block in every step ($m=32$), 25--43\% of the steps of a thread block contain a fallback, and all ants in the block need to wait (Fig.~\ref{fig:kernel_ablation}(b)). In addition, the direct port scans all $N$ cities in each fallback. WCF synchronises only within a warp ($m=4$), which reduces the fraction of waiting steps to 6.5--10.4\%, and the time of each wait is proportional only to the number of unvisited cities. For ACS, the local update synchronises the whole thread block in every step, and 17--39\% of its steps are still waiting steps even with WCF. This explains why ACS has the lowest speedup (Table~\ref{S-tab:ratios}).

\vspace{-3mm}
\section{Conclusions and Future Work}
\label{sec:conclusion}

The goal of this paper was to speed up evaluation in AAD when the algorithm being designed is itself an iterative algorithm, with GP for designing the transition rules of ACO as the case study. The main contributions are as follows. First, we analysed why existing GPU methods are not suitable for this workload. They precompute a fixed rule, and all ants of a batch advance one construction step at a time. Second, we proposed BlockEval, a batch evaluation framework that treats each run of one program on one instance as the unit of parallelism. Its GPU backend executes each run as a whole in one thread block, compiles all programs of a generation into one module in which each program appears only twice, and makes full use of the GPU by executing many runs simultaneously. Third, we proposed the warp-cooperative fallback (WCF), which enumerates the unvisited cities with the threads of an ant group and, for AS and MMAS, synchronises only within a warp, so that a fallback stalls only one warp.

For RQ1, under the default configuration, the GPU backend is 14.6--156 times faster than EvoGP+TensorACO and 11.2--17.8 times faster than EvoGP+\allowbreak TensorACO-CL, and neither of the two is faster than the CPU backend on 32 cores. Even TensorACO with its fixed rule is not faster than the GPU backend, because it scores all cities in every step. In EvoGP+TensorACO-CL, the GPU is idle for 36--46\% of a step for $N\le 200$, waiting for the host to launch the next kernel. The GPU backend makes the same selections in less than a tenth of the time. For RQ2, the GPU backend is 4.1--16.5 times faster than the CPU backend on 32 cores under the default configuration, and it is 40 times faster than executing the runs one at a time on TSP500. The A5000 is fully used with about a thousand tasks per generation, and compiling at every generation takes only 1.6--1.8\,s in the GPU backend. For RQ3, WCF is particularly effective on large problems. The GPU backend is 8.3 times faster than the direct port on TSP500, and the direct port is slower than the CPU backend on TSP1000.

The techniques of BlockEval should also apply to other iterative algorithms with learned components, such as constructive heuristics designed by GP for scheduling and routing, and code generated by LLMs. We will verify these techniques on such algorithms, extend BlockEval to multiple GPUs, and study how more training instances and larger problems affect the learned rules.

\bibliographystyle{splncs04}
\bibliography{refs}

\clearpage
\section*{Supplementary Material}
\setcounter{section}{0}
\setcounter{figure}{0}
\setcounter{table}{0}
\setcounter{equation}{0}
\setcounter{algorithm}{0}
\renewcommand{\thesection}{S\arabic{section}}
\renewcommand{\thefigure}{S\arabic{figure}}
\renewcommand{\thetable}{S\arabic{table}}
\renewcommand{\theequation}{S\arabic{equation}}
\renewcommand{\thealgorithm}{S\arabic{algorithm}}
\renewcommand{\theHsection}{S\arabic{section}}
\renewcommand{\theHfigure}{S\arabic{figure}}
\renewcommand{\theHtable}{S\arabic{table}}
\renewcommand{\theHequation}{S\arabic{equation}}
\renewcommand{\theHalgorithm}{S\arabic{algorithm}}

This supplementary material provides details and additional results that support the main paper.
\begin{itemize}
\item Section~\ref{S-sec:related} discusses further related work and supports the introduction in Section~\ref{sec:intro}.
\item Section~\ref{S-sec:gpaco} gives the function set and terminal set of GP, the training process, the construction kernel and the computation rules of the two backends, and supports Sections~\ref{sec:background} and~\ref{sec:framework}.
\item Section~\ref{S-sec:methods} gives the parameter settings and the details of the compared methods, and supports the experiment design in Section~\ref{sec:setup}.
\item Section~\ref{S-sec:why} reports additional measurements of the existing GPU methods and supports the answer to RQ1 in Section~\ref{sec:results:why}.
\item Section~\ref{S-sec:additional} reports the paired time ratios, all variants of the ablation study, the cost of compiling and the results on other GPUs, and supports the answers to RQ2 and RQ3 in Sections~\ref{sec:results:evogp}, \ref{sec:results:ablation} and~\ref{sec:analysis}.
\item Section~\ref{S-sec:best_tree} shows an example of a learned transition rule.
\end{itemize}

\section{Extended Related Work}
\label{S-sec:related}

This section extends the discussion of related work in Section~\ref{sec:intro} of the main paper.

\paragraph{Evaluation in AAD.} In AAD, a search method such as GP maintains a population of candidate algorithms and improves it generation by generation~\cite{burke2013hh,nguyen2017gphh,zhang2021gpbook}. In each generation, every candidate is run on a set of training instances, and its performance is aggregated into a fitness value. The evaluation time of one generation is the product of the number of candidates, the number of instances and the time of one run. When a candidate is a dispatching rule or a formula, one run is one pass over an instance. When a candidate is an algorithm such as ACO, one run is itself a search of hundreds of iterations, and evaluation dominates the training time. Surrogate models and multi-fidelity evaluation~\cite{zhang2021surrogate,zhang2022multifidelity} have been widely used to reduce the number of individuals that need exact evaluation. Other studies reduce the cost of one run by using small instances, few ants and few iterations~\cite{lin2025memetic,ye2024reevo}. For example, learning-based and LLM-based ACO design methods~\cite{ye2023deepaco,ye2024reevo,kim2025gfacs} compute a heuristic matrix once per instance, run ACO step by step as tensor operations and use a small evaluation budget. These methods save time by approximating the fitness or by reducing the evaluation budget, which may affect the accuracy of the fitness values. In contrast, BlockEval speeds up the exact fitness evaluation, and it can be used together with these methods.

\paragraph{Evolutionary Computation on GPUs.} EvoX~\cite{huang2025evox} and work on tensorised evolutionary multi-objective optimisation~\cite{liang2026tensor} represent populations and operators as tensors and execute them in batches on GPUs. EvoGP~\cite{wu2026evogp} evaluates GP program trees in batches, and a simulation that uses the trees calls it once per step. MetaDE~\cite{chen2025metade} and AutoPSO~\cite{yu2026autopso} run designed DE and PSO variants in batches, and one iteration of these algorithms consists of a few dense tensor operations. These frameworks parallelise across the candidates and leave the control flow of the algorithm on the host. However, each ant of an ACO run makes $N-1$ dependent selections in every iteration, and each selection scores only $K$ candidate cities. Executed in this way, ACO needs at least one kernel launch in every construction step (Fig.~\ref{fig:execution}(c) of the main paper).

\paragraph{ACO on GPUs.} GPU implementations of ACO map one run to the whole GPU. Cecilia et al.~\cite{cecilia2013gpuaco} assigned each ant of AS to a thread block, and their implementation was up to 20 times faster than the sequential one. Del\'{e}vacq et al.~\cite{delevacq2013gpuaco} found this mapping much faster than assigning each ant to a thread in MMAS. Dawson and Stewart~\cite{dawson2013gpuaco} used four warps for each ant of AS and reported a speedup of up to 82 over a CPU implementation. Skinderowicz~\cite{skinderowicz2016gpuacs,skinderowicz2020gpummas} implemented ACS and MMAS with one thread block per ant, candidate lists of 32 cities and as many ants as cities. Cecilia et al.~\cite{cecilia2018gpuaco} further sped up AS with CUDA vectorisation and atomic operations. TensorACO~\cite{yang2024tensoraco} replaces branches with precomputed weights and masks and uses independent roulette wheels to increase parallelism, which changes the selection probabilities. All these studies parallelise one ACO run on one instance. Ayral and Albayrak~\cite{ayral2017gpugp} compiled GP individuals for the GPU and found the compilation cost prohibitive unless it is amortised.

\paragraph{CPUs and GPUs.} A CPU core is designed to complete a single task quickly. It predicts branches, executes instructions out of order and has large caches, which allow each core to follow its own control flow. A GPU trades single-thread performance for throughput~\cite{lee2010debunking}. Its threads are grouped into warps that execute the same instruction, and when the threads of a warp take different branches, some of them are idle~\cite{fung2007warp}. After both platforms were optimised, a GPU was on average only 2.5 times faster than a multi-core CPU over a set of throughput computing kernels~\cite{lee2010debunking}. A GPU need not be faster for computations with many branches and many dependent steps. It is most effective when it executes a very large number of similar threads. The GPU backend of BlockEval executes the runs of a generation as many similar thread blocks and limits the cost of divergent branches to one warp.

\paragraph{Granularity of Parallelism.} Parallel evolutionary algorithms have long been classified by the level at which they are parallelised, from master--slave models that evaluate the fitness in parallel to coarse-grained and fine-grained population models~\cite{alba2002parallelism}. Tensorisation~\cite{liang2026tensor} parallelises at the finest level, within each operation of an algorithm. BlockEval is a master--slave model in which the unit of work is a complete run. On GPUs, a kernel that executes a whole task is known to suffer from divergent control flow, and Laine et al.~\cite{laine2013megakernels} found that executing path tracing stage by stage is faster, partly for this reason. WCF reduces this cost for the fallback of ACO.

\section{GP-ACO and the Two Backends}
\label{S-sec:gpaco}

This section gives the details of GP-ACO and of the two backends that are omitted from Sections~\ref{sec:background} and~\ref{sec:framework} of the main paper.

\subsection{Function Set and Terminal Set}
\label{S-sec:language}

Table~\ref{S-tab:language} lists the function set and the terminal set of GP. Seven of the ten terminals depend on the construction state of the ant. RTau, REta, BaseConf, DistRank and Entropy depend on the feasible set, ConstructProg depends on the construction step, and TurnCos depends on the previous city. The other three do not change within an iteration. As shown in the last two columns, all 20 champion rules, from 10 training runs each on TSP100 and TSP500, read at least one state-dependent terminal, and all of them read BaseConf. In the recorded populations used for the time measurements, 93.9\% of the 25,600 programs read at least one such terminal.

\begin{sidewaystable}
\setlength{\abovecaptionskip}{0pt}\setlength{\belowcaptionskip}{10pt}
\centering
\caption{Function set and terminal set of GP. $i$ is the current city, $h$ the previous city, $j$ a candidate, $C$ the feasible candidate set, $s$ the construction step, $t$ the iteration, $u$ the number of iterations without improvement of the global best, and $\epsilon=10^{-12}$. The abbreviations in brackets are used in the rule expressions. Champion rules is the number of the 20 champion rules (10 training runs each on TSP100 and TSP500) that read the terminal. Programs is the percentage of the 25,600 programs in the recorded populations that read it, after the expressions are simplified by exact identities.}
\label{S-tab:language}
\scriptsize
\begingroup\setlength{\tabcolsep}{3pt}
\begin{tabular}{P{2.8cm}P{8.1cm}P{3.6cm}cc}
\toprule
Name (abbreviation) & Definition & Depends on / range & Champion rules & Programs (\%) \\
\midrule
\multicolumn{5}{P{16.5cm}}{\textit{Function set. Each node output is protected (NaN$\to$0, $\pm\infty\to\pm10$, clipped to $[-10,10]$)}} \\
ADD, SUB, MUL & $a+b$, $a-b$, $ab$ & Binary &  &  \\
PDIV & $ab/(b^{2}+10^{-6})$ & Binary, protected division &  &  \\
MIN, MAX & $\min(a,b)$, $\max(a,b)$ & Binary &  &  \\
ABS, NEG & $|a|$, $-a$ & Unary &  &  \\
\multicolumn{5}{P{16.5cm}}{\textit{Terminals. $S(x_j)=\tanh[(x_j-\bar{x})/(\sigma_x+10^{-8})]$ standardises a value over the feasible candidate set $C$, $b_j=\tau_{ij}\eta_{ij}^{2}$ and $q_j=b_j/\sum_{k\in C}b_k$}} \\
RTau (RT) & $S(\log\max(\tau_{ij},\epsilon))$ & Edge and feasible set, $[-1,1]$ & 4 & 32.4 \\
REta (RE) & $S(\log\max(\eta_{ij},\epsilon))$ & Edge and feasible set, $[-1,1]$ & 10 & 21.1 \\
BaseConf (BC) & $\tanh(\log\max(q_j,\epsilon)+\log|C|)$ & Edge and feasible set, $[-1,1]$ & 20 & 76.4 \\
DistRank (DR) & $1-2r_j/(|C|-1)$, where $r_j$ is the distance rank among the unvisited cities & Edge and feasible set, $[-1,1]$ & 9 & 16.8 \\
Entropy (EN) & $2[-\sum_j q_j\log\max(q_j,\epsilon)]/\log|C|-1$ & Feasible set & 3 & 18.9 \\
ConstructProg (CP) & $2s/(n-1)-1$, where $s$ is the construction step & Construction step & 2 & 17.1 \\
ACOProg (AP) & $2(t-1)/\max(I-1,1)-1$, where $t$ is the iteration & Iteration & 3 & 19.4 \\
Stagnation (ST) & $2\min(u/I,1)-1$, where $u$ is the number of iterations without improvement of the global best & Iteration & 4 & 15.2 \\
MutualRank (MR) & $1-[(r_{ij}-1)+(r_{ji}-1)]/[2\max(n-2,1)]$, clipped to $[0,1]$ & Edge (fixed for an instance) & 3 & 27.0 \\
TurnCos (TC) & $\cos\angle(x_i-x_h,\;x_j-x_i)$, or 0 if there is no previous city & Edge and previous city, $[-1,1]$ & 5 & 12.4 \\
Constants and ERC & $-1,-0.5,0,0.5,1$, ERC drawn from $U[-1,1]$ (FP32) & Scalar &  &  \\
\midrule
\multicolumn{3}{l}{At least one state-dependent terminal} & 20 & 93.9 \\
\bottomrule
\end{tabular}
\endgroup
\end{sidewaystable}

\subsection{Training Process}
\label{S-sec:training}

Algorithm~\ref{S-alg:gpaco} gives the training process of GP. In each generation, $B$ instances are sampled from the training pool, and all $P$ programs are evaluated on them in one batch. Every five generations, the champion is updated on the validation set.

\begin{algorithm}[htbp]
\caption{Training process of GP for designing the transition rules of ACO}
\label{S-alg:gpaco}
\scriptsize
\begin{algorithmic}[1]
\Require Population size $P$, number of generations $G$, number of instances per generation $B$, training pool, validation set
\Ensure Champion rule $p^{\star}$
\State Initialise the population: 10\% are trees that always output 0 (i.e., the original ACO), and the rest are generated by ramped half-and-half
\For{$g=1,\dots,G$}
  \State Sample $B$ instances from the training pool without replacement, and evaluate all $P$ programs in a batch (Section~\ref{sec:framework:structure} of the main paper)
  \If{$g \bmod 5 = 0$}
    \State Evaluate the current top 5 individuals and the champion $p^{\star}$ on the validation set, and update $p^{\star}$ by the validation gap
  \EndIf
  \State Generate the next population by elitism, tournament selection, crossover, mutation and reproduction
\EndFor
\State \Return $p^{\star}$
\end{algorithmic}
\end{algorithm}

\subsection{Construction Kernel on the GPU}
\label{S-sec:kernel}

Algorithm~\ref{S-alg:construct} gives the work of one ant group in one iteration of the construction kernel. It implements the techniques described in Sections~\ref{sec:framework:gpu} and~\ref{sec:framework:fallback} of the main paper.

\begin{algorithm}[htbp]
\caption{One iteration of an ant group in the construction kernel}
\label{S-alg:construct}
\scriptsize
\begin{algorithmic}[1]
\Require task $(p,b)$, iteration $t$, thread $\ell\in\{0,\dots,T-1\}$ of an ant group
\Ensure tours constructed by the ants of this group in iteration $t$
\For{each ant $a$ of this group}
  \State Select the start city of ant $a$ with the random number of counter $(t,a,0)$, and clear its visited bitmap
\EndFor
\For{$s=1,\dots,N-1$}
  \For{each ant $a$ of this group, at its current city $i$}
    \State Thread $\ell$ loads entries $\ell,\ell+T,\ell+2T,\dots$ of the candidate list of city $i$ and masks them with the bitmap \Comment{no branch}
    \If{the masks of all threads are empty} \Comment{fallback to the full graph}
      \State Select $j^{*}$ by WCF (Algorithm~\ref{alg:wcf} of the main paper)
    \Else
      \State Reduce to obtain the mean and variance of each terminal over $C$; thread $\ell$ computes $z_j$ and $w_j$ (Eq.~(\ref{eq:transition}) of the main paper)
      \State Select $j^{*}$ by prefix sums and merging across threads (counter $(t,a,s)$; ACS takes the maximum with probability $q_0$)
    \EndIf
    \State Set bit $j^{*}$ and $i\gets j^{*}$
  \EndFor
  \State ACS performs the local update and synchronises the thread block
\EndFor
\end{algorithmic}
\end{algorithm}

\subsection{Computation Rules of the Two Backends}
\label{S-sec:contract}

In each generation, the evaluator generates code for all programs and prepares the coordinates, nearest-neighbour lists and random-number keys of each instance. It then passes the $P\times B$ tasks to a backend and computes the fitness by Eq.~(\ref{eq:fitness}) of the main paper. Both backends follow the same computation rules. Random numbers are generated by the counter-based generator Philox4x32-10~\cite{salmon2011philox}, whose keys depend only on the root seed, the phase, the generation and the instance, and not on the program. All programs of a generation therefore use the same random numbers on the same instance, and the differences in fitness come only from the rules themselves. In terms of arithmetic, both backends use deterministic single-precision floating point, with FMA contraction and fast-math disabled, and protect the output of each function node in the same way (Table~\ref{S-tab:language}). However, the two backends do not guarantee the same order of floating-point reductions, and over 500 iterations tiny rounding differences grow into different tours. We therefore compare the time only on the same populations and instances (Section~\ref{S-sec:consistency}).

\section{Compared Methods and Settings}
\label{S-sec:methods}

This section gives the parameter settings and the details of the compared methods, and supports the experiment design in Section~\ref{sec:setup} of the main paper. Table~\ref{S-tab:setup} lists all parameters and experimental settings.

\begin{sidewaystable}
\setlength{\abovecaptionskip}{0pt}\setlength{\belowcaptionskip}{10pt}
\centering
\caption{Parameters and experimental settings. In all main experiments, each configuration is run 10 times independently, and each run is a statistical unit.}
\label{S-tab:setup}
\scriptsize
\setlength{\tabcolsep}{3pt}
\begin{tabular}{P{3.4cm}P{7.0cm}P{8.0cm}}
\toprule
Item & Value & Note \\
\midrule
\multicolumn{3}{l}{\textit{GP (DEAP, single tree)}} \\
Population size / generations & $P=100$, $G=50$ (training runs, Section~\ref{S-sec:best_tree}) & 10\% of the initial population are trees that always output 0; the rest are generated by ramped half-and-half (depth 2--4) \\
Tree limits / selection & At most 31 nodes and depth at most 5; tournament (size 4), elitism 10\% & Offspring exceeding the limits are replaced by their parents \\
Crossover / mutation / reproduction & 0.80 / 0.15 / 0.05 & Mutation: subtree 0.5, node replacement 0.3, constant perturbation 0.2 ($\sigma=0.1$) \\
Fitness & Eq.~(\ref{eq:fitness}) of the main paper; $B=32$ training instances resampled in each generation & The champion is updated on the validation set every 5 generations \\
\multicolumn{3}{l}{\textit{ACO (no local search in any variant)}} \\
Ants / iterations / candidate list & $A=32$, $I=500$, $K=20$ & When the candidate list is exhausted, the unvisited city with the largest weight $w_j$ is selected \\
$\alpha$, $\beta$, $\gamma$ & 1, 2, 1/3 & Eq.~(\ref{eq:transition}) of the main paper \\
AS / ACS / MMAS & $\rho=0.5$ / $\rho=0.1$, $\xi=0.1$, $q_0=0.9$ / $\rho=0.02$, $p_{\mathrm{best}}=0.05$ & The local update of ACS is synchronised in every step; MMAS restarts after 250 iterations without improvement \\
\multicolumn{3}{l}{\textit{GPU execution (RTX A5000)}} \\
Thread organisation & One thread block per task: $\min(A,32)$ ant groups $\times$ $T=8$ threads & When $A>32$, the ants are processed in turn \\
Active tasks & 3,200 (training runs); determined by the GPU memory (scalability experiments) & The remaining tasks are executed in batches, keeping 25\% of the GPU memory free \\
Program execution / arithmetic & One CUDA module is generated and compiled per generation; FP32 without FMA contraction or fast-math & Compilation time is included in the evaluation time; the same per-node protection as on the CPU \\
\multicolumn{3}{l}{\textit{Experiments (10 independent runs, a run as the statistical unit)}} \\
Scalability (RQ2) & Three variants; starting point $N=100$, $P=B=A=32$; $P\le256$, $B\le128$, $A\le128$ (powers of 2); $N\in\{50,100,200,500,1000\}$ & Each run evaluates the recorded populations of 10 consecutive generations of a GP run with 256 programs on TSP100 (a configuration with $P$ programs uses $P$ of them in a fixed random order); a $P\times B$ grid and 6 configurations at $N=500$ are also included \\
CPU backend (RQ2) & The same populations and instances as on the GPU; 32 physical cores; additionally $C\in\{1,\dots,16\}$ at $N=100$ & Paired by run; the comparison with the sequential Python implementation uses $P=4$, $B=8$, $A=8$, $C\in\{1,8\}$ \\
TensorACO with its fixed rule (RQ1) & The rule $\tau_{ij}^{\alpha}\eta_{ij}^{\beta}$ of TensorACO, precomputed as a table in every iteration, without programs; all ants deposit pheromone, $\rho=0.5$, $\alpha=1$, $\beta=2$; a workload of the same shape as the default configuration, with 1,024 runs per generation in which each of the 32 instances is solved 32 times; $N\in\{50,100,200,500,1000\}$, 10 runs at each size & All runs of a generation are executed together for $N\le 200$, in batches of 256 runs for $N=500$ and of 64 runs for $N=1000$; only the time is compared (Table~\ref{tab:evogp} of the main paper) \\
EvoGP+TensorACO (RQ1) & The transition weights of TensorACO are modified by the programs, which EvoGP evaluates at every construction step (Eq.~(\ref{eq:transition}) of the main paper), and the rest is unchanged; no candidate lists, each ant selects among all unvisited cities; AS, default configuration, $N\in\{50,100,200,500,1000\}$, at each size the same 10 runs as the two backends in Table~\ref{tab:evogp} of the main paper & The same populations and instances as \textbf{BlockEval (Ours)}; all 1,024 runs of a generation are executed together for $N\le 200$, in batches of 256 runs for $N=500$ and of 32 runs for $N=1000$ (Section~\ref{S-sec:fixed}); only the time is compared \\
EvoGP+TensorACO-CL (RQ1) & EvoGP+TensorACO with candidate lists and the fallback to the full graph; the same candidate lists, selection probabilities and random numbers as \textbf{BlockEval (Ours)}; executed with \texttt{torch.compile}; AS, default configuration, $N\in\{50,100,200,500,1000\}$, at each size the same 10 runs as the two backends in Table~\ref{tab:evogp} of the main paper & The same populations and instances as the two backends of \textbf{BlockEval (Ours)}, paired by run (Table~\ref{tab:evogp} of the main paper); the three execution modes run 3 times at $N\in\{100,500\}$ (Table~\ref{S-tab:compose}); at each size, one more run records the kernel timeline, and its calls to EvoGP are also executed alone (Table~\ref{S-tab:stages}) \\
Ablation: kernels (RQ2, RQ3; Section~\ref{sec:results:ablation} of the main paper) & Direct port: all 111 configurations; unrolled scoring loop: configurations at $N=500$ and 1000 & The same populations and instances as the GPU backend, paired by run; Table~\ref{tab:ablation} of the main paper lists only the default configuration of AS at $N=500$ \\
Ablation: implementation (RQ2, RQ3; Section~\ref{sec:results:ablation} of the main paper) & Mapping: 1/4/8/16 ants per block, and 1 ant per block with one run at a time ($N=500$); interpreter and 4/16 threads per ant group ($N\in\{100,500\}$), execution in batches ($N=500$); AS, default configuration, 3 runs; instrumented runs record the time of each kernel and the number of candidate-list exhaustions & Only medians are reported, without statistical inference; the main paper uses only $N=500$; the mapping variants construct the same tours as the GPU backend \\
\bottomrule
\end{tabular}
\end{sidewaystable}

\subsection{Speed of the CPU Backend}
\label{S-sec:cpu_baseline}

\begin{figure}[htbp]
\centering
\begin{minipage}[t]{0.34\textwidth}
\centering
\includegraphics[width=\linewidth]{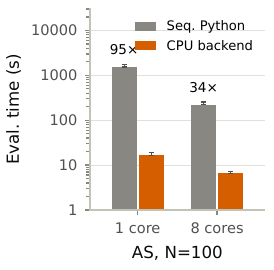}\\[2pt]
\vspace{-1mm}
{\small (a) Sequential Python vs.\ the CPU backend}
\end{minipage}\hfill
\begin{minipage}[t]{0.54\textwidth}
\centering
\includegraphics[width=\linewidth]{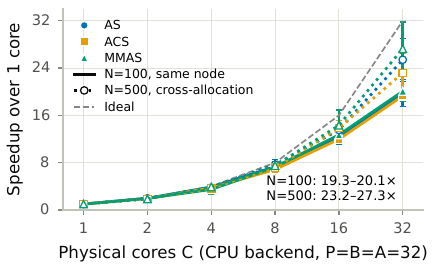}\\[2pt]
\vspace{-1mm}
{\small (b) Core scaling of the CPU backend}
\end{minipage}
\vspace{-2mm}
\caption{(a) Evaluation time per generation of the sequential Python implementation and the CPU backend (AS, $N=100$, $P=4$, $B=8$, $A=8$), and their ratios. (b) Speedup of the CPU backend over a single core ($P=B=A=32$) on TSP100 (solid) and TSP500 (dotted), with the values at 32 cores. The grey dashed line shows linear speedup. Error bars show the standard deviation.}
\label{S-fig:cpu_baseline}
\end{figure}

The CPU backend is much faster than the sequential Python implementation used in existing GP-ACO studies~\cite{lin2025memetic} (Fig.~\ref{S-fig:cpu_baseline}(a)). On one core, it is 95 times faster than this implementation, and on eight cores it is 34 times faster, because every selection of the Python implementation goes through the interpreter. Fig.~\ref{S-fig:cpu_baseline}(b) shows the speedup of the CPU backend over a single core. On TSP100, 32 cores are 19.3--20.1 times faster than one core.

\subsection{TensorACO and EvoGP+TensorACO}
\label{S-sec:fixed}

TensorACO~\cite{yang2024tensoraco} represents the ants of a run as tensors. At the beginning of an iteration, it computes the transition weights $\tau_{ij}^{\alpha}\eta_{ij}^{\beta}$ of all edges as an $N\times N$ table. In each construction step, each ant reads the row of its current city from this table, masks the visited cities and selects the next city with the adaptive independent roulette of TensorACO. There are no candidate lists, and each ant selects among all unvisited cities in every step. After the last step, the pheromone is updated from the tours of all ants.

We run TensorACO with its own fixed rule on a workload of the same shape as the default configuration. A generation consists of 1,024 runs with 32 ants and 500 iterations, in which each of the 32 instances is solved 32 times with different random numbers. Its released code processes one run at a time. We stack the runs of a generation along a new axis of the tensors and execute them together, in batches of the size given in Table~\ref{S-tab:nocl}. For $N\le 200$, stacking makes the released code 28--88 times faster.

EvoGP+TensorACO keeps the procedure of TensorACO and changes only the rule. In each construction step, the terminals of Table~\ref{S-tab:language} are computed for all cities of each ant, with all unvisited cities as the feasible set. EvoGP then evaluates the program of each run on these rows, and the weight of each city is multiplied by $1+\gamma\tanh(z_j)$ as in Eq.~(\ref{eq:transition}) of the main paper. The selection and the pheromone update are those of TensorACO, and the runs are stacked in the same way. Since its feasible set and selection differ from those of BlockEval, we compare only the evaluation time.

In long runs, both methods return tours that visit a city more than once, because the pheromone of rarely used edges underflows in single precision. For TensorACO, the share of such tours in the last iteration rises from 1.3\% at $N=50$ to 95.2\% at $N=1000$ (Table~\ref{S-tab:nocl}). This does not change the amount of computation in a step.

In Table~\ref{tab:evogp} of the main paper, the GPU memory is the peak memory held by the process, sampled during the run. TensorACO is written in JAX, which reserves GPU memory in blocks and by default uses at most 75\% of the available memory, or 17.66\,GB on the A5000. TensorACO reaches this limit from $N=200$. Its pheromone update builds an $N\times N$ matrix for each ant, which takes $4AN^2$ bytes per run. For the 1,024 runs of a generation, these matrices alone exceed the 24\,GB of the A5000 when $N\ge 500$, and the runs are executed in batches in turn. EvoGP+TensorACO also stores one row of terminals for each city of each ant and needs 20.6\,GB at $N=200$ (Table~\ref{tab:evogp} of the main paper).

\begin{table}[!htb]
\centering
\caption{The two methods without candidate lists under the default configuration of AS. As released is the time per generation when the released code of TensorACO executes one run at a time, measured in the first generation of three runs for $N\le 200$ only. The speedup of stacking is this time divided by the time with the runs stacked in Table~\ref{tab:evogp} of the main paper. Revisiting tours is the percentage of tours in the last iteration that visit a city more than once. Time / fixed rule is the time per generation of EvoGP+TensorACO divided by that of TensorACO with its fixed rule.}
\label{S-tab:nocl}
\scriptsize
\begingroup\setlength{\tabcolsep}{4pt}
\begin{tabular}{cccccccc}
\toprule
 & \multicolumn{4}{c}{TensorACO (fixed rule)} & \multicolumn{3}{c}{EvoGP+TensorACO} \\
\cmidrule(lr){2-5}\cmidrule(lr){6-8}
$N$ & \shortstack[c]{As\\released\\(s)} & \shortstack[c]{Speedup\\of\\stacking} & \shortstack[c]{Runs\\per\\batch} & \shortstack[c]{Revisiting\\tours\\(\%)} & \shortstack[c]{Runs\\per\\batch} & \shortstack[c]{Time /\\fixed\\rule} & \shortstack[c]{Revisiting\\tours\\(\%)} \\
\midrule
50 & 283.2 & 88 & 1,024 & 1.3 & 1,024 & 14.0 & 1.3 \\
100 & 552.6 & 57 & 1,024 & 4.6 & 1,024 & 17.8 & 4.5 \\
200 & 948.1 & 28 & 1,024 & 15.2 & 1,024 & 20.5 & 14.9 \\
500 & -- & -- & 256 & 62.1 & 256 & 17.9 & 61.8 \\
1000 & -- & -- & 64 & 95.2 & 32 & 15.8 & 95.1 \\
\bottomrule
\end{tabular}
\endgroup
\end{table}

\subsection{EvoGP+TensorACO-CL}
\label{S-sec:cl}

EvoGP+TensorACO-CL adds candidate lists and the fallback to the full graph to EvoGP+TensorACO. Its candidate lists, terminals, transition rule, roulette wheel, fallback rule, pheromone update and random numbers are the same as in the two backends of BlockEval, and it evaluates the same workload (Section~\ref{S-sec:consistency}). All $P\times B\times A$ ants advance one construction step together. In each step, the terminals of the $K$ candidates of all ants are computed as tensors, EvoGP evaluates the programs on these $PBAK$ rows, and the roulette wheel selects the next city. Next, the host waits until the GPU reports how many ants have exhausted their candidate lists, and these ants are gathered. Since all ants are at the same step, these ants have the same number of unvisited cities. Their terminals are computed on a compact array of these cities, followed by a second call to EvoGP. After each iteration, the pheromone is updated ant by ant, as in the two backends.

EvoGP evaluates each row of inputs with one thread, which interprets the tree stored with the row. We added the protected division and the protection of every node output (Table~\ref{S-tab:language}) to its tree-evaluation kernel. Calling EvoGP at every construction step is consistent with its use in control problems, where it is called once per simulation step.

We execute the tensor operations in three modes. The eager mode executes them one by one, the CUDA graph mode captures the ordinary selection and the update of the tours as CUDA graphs, and the \texttt{torch.compile} mode compiles them. As shown in Table~\ref{S-tab:compose}, the \texttt{torch.compile} mode is the fastest of the three, and it is the one reported in the main paper. As in Table~\ref{tab:evogp} of the main paper, the times exclude the one-time cost of the first generation, which is the compilation of \texttt{torch.compile} or the warm-up of the other two modes. CUDA graphs reduce the eager time by only 11--13\%, since the host still waits for the GPU in every step before it launches the fallback.

\begin{table}[!htb]
\centering
\caption{Evaluation time per generation of EvoGP+TensorACO-CL in three execution modes (AS, $P=B=A=32$, RTX A5000). The values are medians over three runs and exclude the one-time cost of the first generation. The ratio is the time divided by that of the GPU backend of BlockEval, paired by run.}
\label{S-tab:compose}
\scriptsize
\setlength{\tabcolsep}{5pt}
\begin{tabular}{ccccc}
\toprule
Execution & $N=100$ (s) & Ratio & $N=500$ (s) & Ratio \\
\midrule
Eager & 188.4 & 37.7 & 1016.3 & 24.4 \\
CUDA graph & 163.5 & 32.1 & 906.4 & 22.4 \\
\texttt{torch.compile} & 91.93 & 18.4 & 512.3 & 12.8 \\
\midrule
\textbf{BlockEval (Ours)}, GPU backend & \textbf{4.99} & 1 & \textbf{40.37} & 1 \\
\bottomrule
\end{tabular}
\end{table}

\subsection{Workload Consistency}
\label{S-sec:consistency}

Table~\ref{S-tab:consistency} compares the tours and fitness values of the implementations on the same workload. The two backends construct identical tours in 5.5--70.7\% of the tasks for AS and in 66.7--99.6\% for ACS and MMAS, and the tours of the other tasks differ because of rounding (Section~\ref{S-sec:contract}). Over all 111 configurations, the average gap of the same program differs between the two backends by 0.005 percentage points on average and by at most 0.23. EvoGP+TensorACO-CL constructs identical tours in about the same share of tasks as the CPU backend for AS. Its total number of fallbacks differs from that of the GPU backend by less than 0.05\%, and the mean difference in the average gap of the same program is at most 0.12 percentage points at each size. For a single program, the difference is at most 1.4 percentage points. On small instances, its three execution modes construct the same tours as the CPU backend.

\begin{table}[!htb]
\centering
\caption{Consistency of the implementations on the same workload under the default configuration ($P=B=A=32$). Identical tours is the percentage of tasks for which the CPU backend or EvoGP+TensorACO-CL constructs exactly the same tours as the GPU backend. For EvoGP+TensorACO-CL, which runs AS on the 10 runs of Table~\ref{tab:evogp} of the main paper, the ratio of fallbacks is its total number of fallbacks divided by that of the GPU backend. The gap difference is the average gap of the same program minus that in the GPU backend, in percentage points, given as the mean over all programs and as the largest absolute value for a single program. The last row gives the mean and the largest absolute value of the gap difference between the two backends (GPU minus CPU) over all 111 configurations.}
\label{S-tab:consistency}
\scriptsize
\begingroup\setlength{\tabcolsep}{3pt}
\begin{tabular}{cccccccc}
\toprule
 & \multicolumn{3}{c}{CPU backend, identical tours (\%)} & \multicolumn{4}{c}{EvoGP+TensorACO-CL (AS)} \\
\cmidrule(lr){2-4}\cmidrule(lr){5-8}
$N$ & AS & ACS & MMAS & \shortstack{Identical\\tours (\%)} & \shortstack{Ratio of\\fallbacks} & \shortstack{Mean gap\\difference} & \shortstack{Largest for\\a program} \\
\midrule
50 & 70.7 & 99.6 & 99.6 & 69.8 & 1.0001 & 0.002 & 0.34 \\
100 & 11.4 & 98.9 & 98.1 & 11.3 & 1.0000 & 0.005 & 0.98 \\
200 & 5.5 & 94.8 & 95.2 & 5.5 & 1.0000 & 0.000 & 1.36 \\
500 & 7.7 & 82.0 & 85.6 & 7.7 & 0.9998 & $-$0.043 & 1.08 \\
1000 & 34.0 & 66.7 & 72.0 & 32.4 & 0.9996 & $-$0.116 & 1.15 \\
\midrule
\multicolumn{8}{P{9cm}}{Gap difference between the two backends over all 111 configurations, mean 0.005, largest absolute value 0.23} \\
\bottomrule
\end{tabular}
\endgroup
\end{table}

\subsection{Conditions of the Comparison}
\label{S-sec:conditions}

All methods run on the same RTX A5000, with no other process on the GPU. EvoGP+TensorACO and EvoGP+TensorACO-CL evaluate the same recorded populations and instances as the two backends of BlockEval, and EvoGP+\allowbreak TensorACO-CL also uses the same random-number keys and is paired by run. The existing GPU methods are compiled once, and this one-time compilation is excluded from their time per generation. For EvoGP+TensorACO-CL, it takes 33.2--39.8\,s per problem size (Table~\ref{S-tab:compile}). Both backends of BlockEval compile at every generation, and their times include this compilation.

\section{Why Existing GPU Methods Are Slower}
\label{S-sec:why}

This section reports additional measurements of the existing GPU methods and supports the answer to RQ1 in Section~\ref{sec:results:why} of the main paper.

\subsection{Without Candidate Lists}
\label{S-sec:why:nocl}

Without candidate lists, each of the $PBA=32{,}768$ ants scores all $N$ cities in every step. For TensorACO and EvoGP+TensorACO, the time per construction step therefore grows in proportion to $N$, and the time per generation grows with $N^2$ (Tables~\ref{tab:evogp} and~\ref{tab:steps} of the main paper). Although the two methods differ only in the rule, EvoGP+TensorACO takes 14.0--20.5 times longer than TensorACO with its fixed rule (Table~\ref{S-tab:nocl}).

To investigate how much of this time is spent on computing the programs, we relate the time of EvoGP+TensorACO to the size of its programs (Table~\ref{S-tab:nodes}). The programs of the recorded populations have 1--31 nodes. At $N=1000$, a batch contains the 32 runs of one program. A batch whose program is a single terminal requires no arithmetic but still needs 23.5\,ms per step, against 59.9\,ms for programs with 25--31 nodes. At all five problem sizes, the time of a batch grows linearly with the mean number of nodes of its programs ($R^2\ge 0.95$). The intercept of the fit does not depend on the program and accounts for 60--64\% of the time at the mean program size. This indicates that most of the time of EvoGP+TensorACO is spent on computing the terminals of all cities, passing them to EvoGP, returning the scores and selecting the next city, rather than on computing the programs.

\begin{table}[!htb]
\centering
\caption{Time per construction step of EvoGP+TensorACO against the size of the programs. The upper part groups the batches at $N=1000$ by the number of nodes of their program. At this size, a batch contains the 32 runs of one program, and the time of a batch is converted to the time in which all 1,024 runs advance one step. The lower part gives a linear fit of the time of a batch to the mean number of nodes of its programs at each problem size, with the intercept, which does not depend on the program, the slope per node and the share of the intercept at the mean program size.}
\label{S-tab:nodes}
\scriptsize
\begingroup\setlength{\tabcolsep}{5pt}
\begin{tabular}{ccc}
\toprule
Nodes of the program & Batches & Time per step (ms) \\
\midrule
1 & 82 & 23.5 \\
2--4 & 315 & 26.9 \\
5--8 & 1096 & 30.4 \\
9--12 & 815 & 35.6 \\
13--16 & 520 & 40.8 \\
17--24 & 282 & 49.8 \\
25--31 & 90 & 59.9 \\
\bottomrule
\end{tabular}
\endgroup

\vspace{2mm}

\begingroup\setlength{\tabcolsep}{5pt}
\begin{tabular}{lccccc}
\toprule
Linear fit & $N=50$ & 100 & 200 & 500 & 1000 \\
\midrule
Intercept (ms) & 1.17 & 2.12 & 4.19 & 11.3 & 21.7 \\
Slope per node (ms) & 0.065 & 0.136 & 0.277 & 0.661 & 1.364 \\
Share of the intercept (\%) & 64.1 & 60.7 & 60.1 & 63.0 & 61.3 \\
$R^2$ & 0.95 & 0.97 & 0.98 & 0.97 & 0.96 \\
\bottomrule
\end{tabular}
\endgroup
\end{table}

\subsection{With Candidate Lists}
\label{S-sec:why:cl}

For EvoGP+TensorACO-CL, the GPU is idle, waiting for the host to launch the next kernel, for 37--43\% of the time for $N\le 200$, 21\% at $N=500$ and 7\% at $N=1000$ (upper part of Table~\ref{S-tab:stages}). For TensorACO and EvoGP+TensorACO, this share is at most 5.3\%, and their time is spent in kernels. For the GPU backend, the GPU is idle only while the host compiles the programs, and this share differs from the share of the compilation measured by the timers by at most 0.6 percentage points.

To measure the kernel time of each stage, we record the start and the end of every kernel in 10 evenly sampled iterations of one generation of EvoGP+\allowbreak TensorACO-CL at each problem size (middle part of Table~\ref{S-tab:stages}). The kernels run for 1.00--2.33\,ms per step, and the GPU is idle for the remaining 0.26--0.90\,ms, which agrees with the upper part. Of the kernel time, EvoGP takes 0.49--0.50\,ms per step regardless of $N$, and computing the terminals takes 0.19--0.23\,ms. The kernel time of the fallback grows with $N$, from 0.10\,ms to 1.34\,ms.

A step launches 67--75 kernels, and the host waits for the GPU about once per step, when four bytes are copied to the host (lower part of Table~\ref{S-tab:stages}). When the $I(N-1)$ calls to EvoGP of a generation are executed alone, a call takes 0.46\,ms, which is longer than a complete step of the GPU backend (Table~\ref{tab:steps} of the main paper). Each of the 655,360 rows of a call carries its own copy of the program in 32 node slots of 8 bytes. This copy takes 256 bytes, whereas the 10 single-precision terminals of the row take 40 bytes.

\begin{table}[!htb]
\centering
\caption{A construction step of EvoGP+TensorACO-CL under the default configuration of AS ($P=B=A=32$, \texttt{torch.compile} mode, RTX A5000). The upper part is the share of the time in which the GPU is idle, waiting for the host to launch the next kernel, in the runs of Table~\ref{tab:evogp} of the main paper. The driver reports once per second the share of the last period in which at least one kernel was running, and the table gives its complement, averaged over the measured generations of each run and then over the runs. The last row of this part is the share of the compilation in the evaluation time of the GPU backend, measured by the timers. The middle part is the kernel time per construction step by stage, from one generation at each problem size in which every kernel is recorded in 10 evenly sampled iterations. The time per step of this generation is taken from the same generation of the runs of Table~\ref{tab:evogp}, and its difference to the time of all kernels is the time in which the GPU is idle. The lower part gives the kernels launched, the waits of the host and the bytes copied to the host per step, and the time per step when only the calls to EvoGP of a generation are executed (means over 10 runs). }
\label{S-tab:stages}
\scriptsize
\begingroup\setlength{\tabcolsep}{3.5pt}
\begin{tabular}{lccccc}
\toprule
 & $N=50$ & 100 & 200 & 500 & 1000 \\
\midrule
\multicolumn{6}{l}{GPU idle, waiting for the host to launch the next kernel, in the runs of Table~\ref{tab:evogp} (\%)} \\
\quad TensorACO (fixed rule) & 5.3 & 2.8 & 1.2 & 1.4 & 0.5 \\
\quad EvoGP+TensorACO & 0.6 & 0.3 & 0.0 & 0.0 & 0.1 \\
\quad EvoGP+TensorACO-CL & 43.1 & 43.1 & 37.4 & 20.7 & 7.2 \\
\quad \textbf{BlockEval (Ours)}, GPU backend & 57.9 & 35.0 & 13.7 & 4.3 & 1.4 \\
\quad Share of compilation, GPU backend (timers) & 57.5 & 34.9 & 14.3 & 4.1 & 1.4 \\
\midrule
\multicolumn{6}{l}{Kernel time per construction step of EvoGP+TensorACO-CL (ms)} \\
\quad Computing the terminals & 0.193 & 0.200 & 0.209 & 0.218 & 0.231 \\
\quad EvoGP in the ordinary selection & 0.503 & 0.499 & 0.489 & 0.488 & 0.492 \\
\quad Roulette wheel & 0.047 & 0.053 & 0.067 & 0.092 & 0.107 \\
\quad Fallback to the full graph & 0.101 & 0.170 & 0.295 & 0.650 & 1.337 \\
\quad \quad of which calls to EvoGP & 0.014 & 0.023 & 0.036 & 0.084 & 0.161 \\
\quad Writing the tours & 0.026 & 0.027 & 0.028 & 0.030 & 0.031 \\
\quad Candidate tables and pheromone update & 0.127 & 0.099 & 0.098 & 0.112 & 0.131 \\
\quad All kernels & 0.998 & 1.048 & 1.186 & 1.589 & 2.329 \\
\quad Time per step of this generation & 1.86 & 1.94 & 1.86 & 2.04 & 2.59 \\
\quad GPU idle, waiting for the host to launch the next kernel & 0.863 & 0.896 & 0.678 & 0.456 & 0.261 \\
\midrule
Kernels per step & 75 & 70 & 68 & 67 & 67 \\
Host waits per step & 1.06 & 1.03 & 1.02 & 1.01 & 1.00 \\
Bytes copied to the host per step & 4 & 4 & 4 & 4 & 4 \\
Its calls to EvoGP alone (ms per step) & 0.461 & 0.462 & 0.462 & 0.463 & 0.462 \\
\bottomrule
\end{tabular}
\endgroup
\end{table}

\subsection{Comparison with the CPU Backend}
\label{S-sec:why:cpu}

Fig.~\ref{fig:scaling}(c) of the main paper shows that EvoGP+TensorACO-CL comes closer to the CPU backend as the number of runs per generation increases. Its time grows slowly with the number of runs, whereas that of the CPU backend grows roughly linearly. Its throughput reaches that of the CPU backend at about 2,048 runs per generation. However, the GPU backend is still at least 8.8 times faster than EvoGP+TensorACO-CL at all points of Fig.~\ref{fig:scaling} (Table~\ref{S-tab:ratios}).

\section{Additional Results of BlockEval}
\label{S-sec:additional}

This section supports the answers to RQ2 and RQ3 in Sections~\ref{sec:results:evogp}, \ref{sec:results:ablation} and~\ref{sec:analysis} of the main paper.

\subsection{Paired Time Ratios}
\label{S-sec:ratios}

Table~\ref{S-tab:ratios} gives the evaluation time per generation of each method divided by that of the GPU backend, paired by run. The times of AS are those in Table~\ref{tab:evogp} of the main paper, and those of ACS and MMAS are plotted in Fig.~\ref{fig:scaling}(a). For all three variants, the speedup of the GPU backend over the CPU backend is highest at $N=100$ and decreases as $N$ increases further. ACS has the lowest speedup at all five problem sizes, because its local update synchronises the whole thread block in every step (Section~\ref{sec:analysis:wcf} of the main paper). The last row gives the ranges over all configurations of Fig.~\ref{fig:scaling}. Over the 111 configurations, the GPU backend is 1.70--18.0 times faster than the CPU backend, and at the 23 points of the grey line, it is 8.8--55.6 times faster than EvoGP+TensorACO-CL.

\begin{table}[!htb]
\centering
\caption{Evaluation time per generation of each method divided by that of the GPU backend of BlockEval under the default configuration ($P=B=A=32$). The ratios are paired by run and averaged over 10 runs, and the ratios of the CPU backend on 32 cores are the speedups of the GPU backend. The last row gives the ranges over all configurations of Fig.~\ref{fig:scaling} of the main paper, with medians over three runs for EvoGP+TensorACO-CL. }
\label{S-tab:ratios}
\scriptsize
\begingroup\setlength{\tabcolsep}{3.5pt}
\begin{tabular}{ccccccc}
\toprule
 & \multicolumn{4}{c}{AS} & ACS & MMAS \\
\cmidrule(lr){2-5}\cmidrule(lr){6-6}\cmidrule(lr){7-7}
$N$ & \shortstack{TensorACO\\(fixed rule)} & \shortstack{EvoGP+\\TensorACO} & \shortstack{EvoGP+\\TensorACO-CL} & \shortstack{CPU backend\\(32 cores)} & \shortstack{CPU backend\\(32 cores)} & \shortstack{CPU backend\\(32 cores)} \\
\midrule
50 & 1.05 & 14.61 & 14.45 & 14.14 & 10.76 & 13.94 \\
100 & 1.85 & 33.04 & 17.76 & 14.78 & 11.80 & 16.52 \\
200 & 3.01 & 61.65 & 17.09 & 11.45 & 8.95 & 13.94 \\
500 & 6.34 & 113.5 & 13.04 & 8.88 & 6.14 & 10.20 \\
1000 & 9.86 & 156.0 & 11.16 & 7.39 & 4.08 & 8.19 \\
\midrule
\multicolumn{3}{l}{All configurations of Fig.~\ref{fig:scaling}} & 8.8--55.6 & \multicolumn{3}{c}{1.70--18.0} \\
\bottomrule
\end{tabular}
\endgroup
\end{table}

\subsection{Ablation Study: All Variants}
\label{S-sec:ablation}

Table~\ref{S-tab:ablation} lists all variants of the ablation study, of which Table~\ref{tab:ablation} of the main paper shows six. Each variant changes one design choice of the GPU backend on TSP500 under the default configuration of AS. In the first group, the tasks are not all executed at once. When the runs are executed one at a time with one ant per thread block, as in existing GPU implementations of ACO~\cite{delevacq2013gpuaco,skinderowicz2020gpummas}, each kernel launch uses only 32 thread blocks of 8 threads. These blocks occupy at most 32 of the 64 SMs of the A5000, and the 1,024 runs of a generation are executed one after another. The GPU backend is 40.0 times faster than this variant. Batches of 256 or 512 tasks take 1.22 and 1.07 times as long as executing all tasks at once, respectively.

In the second group, the ants of a task are spread over several thread blocks. A block executes $A'\in\{1,4,8,16\}$ ants of one task with $8A'$ threads, a task occupies $A/A'$ consecutive blocks, and the arrays in shared memory are allocated for the ants of the block only. The candidate lists, terminals, selection, fallback, pheromone update and random numbers are unchanged, and these variants construct the same tours as the GPU backend at every generation. They still take 1.05--1.70 times as long. The GPU backend asks the compiler to keep three resident blocks per SM, whereas the smaller blocks are compiled without this requirement and use more registers per thread. Fewer ants can then reside on an SM. For example, at most 16 blocks of one ant reside on an SM, against three blocks of 32 ants in the GPU backend, and with one ant per block, three quarters of each warp are empty. This mapping is not available for ACS, whose local update synchronises all ants of a task after every step.

The direct port scans all $N$ cities in the fallback, computes DistRank by pairwise comparisons and synchronises the whole thread block in every step. Its cost per transition and its fraction of waiting steps are shown in Fig.~\ref{fig:kernel_ablation} of the main paper. The unrolled scoring loop uses up to 128 registers per thread, against at most 80 in the GPU backend, which reduces the resident tasks per SM from three to two in most generations. With 16 threads per ant group, most threads are idle during the ordinary selection. All variants follow the same computation rules as the GPU backend. These results indicate that the speedup of the GPU backend comes from how the runs are executed rather than from details of the implementation.

\begin{table}[!htb]
\centering
\caption{All variants of the ablation study on TSP500 (AS, $P=B=A=32$, RTX A5000). The time ratio is the evaluation time per generation of the variant divided by that of the GPU backend, as the mean over paired runs for the direct port and the unrolled scoring loop, and as the median over three runs for the others. The last three columns give the number of threads per thread block, the number of registers per thread and the number of thread blocks that can reside on an SM for the construction kernel of each variant. A range is given when the number of registers changes with the programs of a generation.}
\label{S-tab:ablation}
\scriptsize
\begingroup\setlength{\tabcolsep}{2.5pt}
\begin{tabular}{llcccc}
\toprule
Design choice & Variant & Time ratio & \shortstack{Threads\\per block} & \shortstack{Registers\\per thread} & \shortstack{Blocks\\per SM} \\
\midrule
GPU backend & -- & 1 & 256 & 78--80 & 3 \\
\midrule
All tasks at once & One run at a time & 40.0 & 8 & 128--159 & 12--16 \\
 & Batches of 256 tasks & 1.22 & 256 & 80 & 3 \\
 & Batches of 512 tasks & 1.07 & 256 & 80 & 3 \\
\midrule
One task per block & 1 ant per block & 1.70 & 8 & 128--159 & 12--16 \\
 & 4 ants per block & 1.19 & 32 & 115--128 & 16 \\
 & 8 ants per block & 1.11 & 64 & 94--116 & 8--10 \\
 & 16 ants per block & 1.05 & 128 & 80--96 & 5--6 \\
\midrule
WCF & Direct port & 8.30 & 256 & 64--94 & 2--4 \\
\midrule
Compiling once & Unrolled scoring loop & 1.15 & 256 & 80--128 & 2--3 \\
 & Interpreter & 1.38 & 256 & 80 & 3 \\
\midrule
8 threads per ant group & 4 threads per ant group & 1.60 & 128 & 146--168 & 3 \\
 & 16 threads per ant group & 1.34 & 512 & 40 & 3 \\
\bottomrule
\end{tabular}
\endgroup
\end{table}

\subsection{Cost of Compiling at Every Generation}
\label{S-sec:compile}

The CPU backend compiles one scoring function for each program structure and needs 15.8--18.0\,s per generation under the default configuration of AS (Table~\ref{S-tab:compile}). This is 37.0\% of its evaluation time at $N=50$ and 2.0\% at $N=1000$. If the compilation is excluded from the CPU time only, the GPU backend is still 7.24--11.33 times faster. The compilation time also grows with the number of programs and with the number of copies of each program in the kernel. At $P=256$, the direct port with one copy compiles in 4.19\,s, the GPU backend with two copies in 10.49\,s and the unrolled scoring loop with five copies in 34.97\,s.

\begin{table}[!htb]
\centering
\caption{Compilation time under AS. The upper part gives the one-time compilation of EvoGP+TensorACO-CL by \texttt{torch.compile}, the compilation time per generation of the CPU backend on 32 cores, its share of the evaluation time, and the speedup of the GPU backend when the compilation is excluded from the CPU time only ($P=B=A=32$, means over 10 runs). The compilation time of the GPU backend is given in Table~\ref{tab:evogp} of the main paper. The lower part gives the number of copies of each program in three kernels and their compilation time per generation at $P=1$, 32 and 256, as means over all configurations with the same $P$.}
\label{S-tab:compile}
\scriptsize
\begingroup\setlength{\tabcolsep}{5pt}
\begin{tabular}{ccccc}
\toprule
$N$ & \shortstack{EvoGP+TensorACO-CL\\compiled once (s)} & \shortstack{CPU backend\\compile (s)} & \shortstack{CPU backend\\compile share (\%)} & \shortstack{Speedup without\\CPU compilation} \\
\midrule
50 & 33.2 & 16.0 & 37.0 & 8.92 \\
100 & 34.5 & 18.0 & 23.4 & 11.33 \\
200 & 37.4 & 15.8 & 12.3 & 10.05 \\
500 & 35.8 & 16.9 & 4.8 & 8.46 \\
1000 & 39.8 & 16.8 & 2.0 & 7.24 \\
\bottomrule
\end{tabular}
\endgroup

\vspace{2mm}

\begingroup\setlength{\tabcolsep}{6pt}
\begin{tabular}{lcccc}
\toprule
Kernel & Copies & $P=1$ (s) & $P=32$ (s) & $P=256$ (s) \\
\midrule
Direct port & 1 & 0.66 & 1.09 & 4.19 \\
\textbf{BlockEval (Ours)} & 2 & 0.88 & 1.65 & 10.49 \\
Unrolled scoring loop & 5 & 0.82 & 2.86 & 34.97 \\
\bottomrule
\end{tabular}
\endgroup
\end{table}

\subsection{Results on Other GPUs}
\label{S-sec:gpus}

Table~\ref{S-tab:cross_gpu} gives the speed of six other GPUs relative to the A5000 on the same workload, measured by the device time. The speed grows mainly with the number of SMs, on each of which three tasks reside. The A40 and RTX A6000 have 1.3 times as many SMs as the A5000 and are 1.11--1.26 times faster. The L40S and RTX 6000 Ada have 2.2 times as many SMs and are 1.97--2.64 times faster, and the L4, which has fewer SMs, is slower. The L2 cache and the memory bandwidth affect the speed at different problem sizes. At $N=100$, the pheromone matrix of a task takes $4N^2$ bytes, or 40\,KB, and the 48--96\,MB L2 cache of the Ada GPUs holds the pheromone of all resident tasks. This is consistent with their largest advantage at $N=100$. At $N=1000$, each task needs 4\,MB, and the bandwidth has a larger effect. The relative speed of the L4, with 300\,GB/s, drops from 0.92 to 0.74, whereas that of the RTX PRO 5000, with 1,344\,GB/s, rises from 1.30 to 1.46. The RTX PRO 5000 has the same L2 cache as the L40S and RTX 6000 Ada but is only as fast as the RTX A6000 at $N=100$, which suggests that the architecture itself also has an effect.

\begin{table}[!htb]
\centering
\caption{Speed of each GPU relative to the RTX A5000 under the default configuration ($P=B=A=32$), as medians over the three variants, sorted by the speed at $N=1000$. The speed is the device time on the A5000 divided by that on the GPU, and a value greater than 1 means faster. Arch.\ is the architecture and BW the memory bandwidth. The numbers of SMs, the L2 cache and the bandwidth are obtained from the CUDA device properties.}
\label{S-tab:cross_gpu}
\scriptsize
\setlength{\tabcolsep}{5pt}
\begin{tabular}{cccccccc}
\toprule
\multirow{2}{*}{GPU} & \multirow{2}{*}{Arch.} & \multirow{2}{*}{SMs} & \multirow{2}{*}{L2 (MB)} & \multirow{2}{*}{\shortstack{BW\\(GB/s)}} & \multicolumn{3}{c}{Relative speed} \\
\cmidrule(lr){6-8}
 & & & & & $N{=}100$ & $N{=}500$ & $N{=}1000$ \\
\midrule
L4 & Ada & 58 & 48 & 300 & 0.92 & 0.83 & 0.74 \\
RTX A5000 & Ampere & 64 & 6 & 768 & 1.00 & 1.00 & 1.00 \\
A40 & Ampere & 84 & 6 & 696 & 1.12 & 1.13 & 1.11 \\
RTX A6000 & Ampere & 84 & 6 & 768 & 1.26 & 1.20 & 1.21 \\
RTX PRO 5000 & Blackwell & 110 & 96 & 1,344 & 1.30 & 1.30 & 1.46 \\
L40S & Ada & 142 & 96 & 864 & 2.62 & 2.07 & 1.97 \\
RTX 6000 Ada & Ada & 142 & 96 & 960 & 2.64 & 2.33 & 1.97 \\
\bottomrule
\end{tabular}
\end{table}

\section{An Example of a Learned Transition Rule}
\label{S-sec:best_tree}

\begin{figure}[htbp]
\centering
\includegraphics[width=0.8\textwidth]{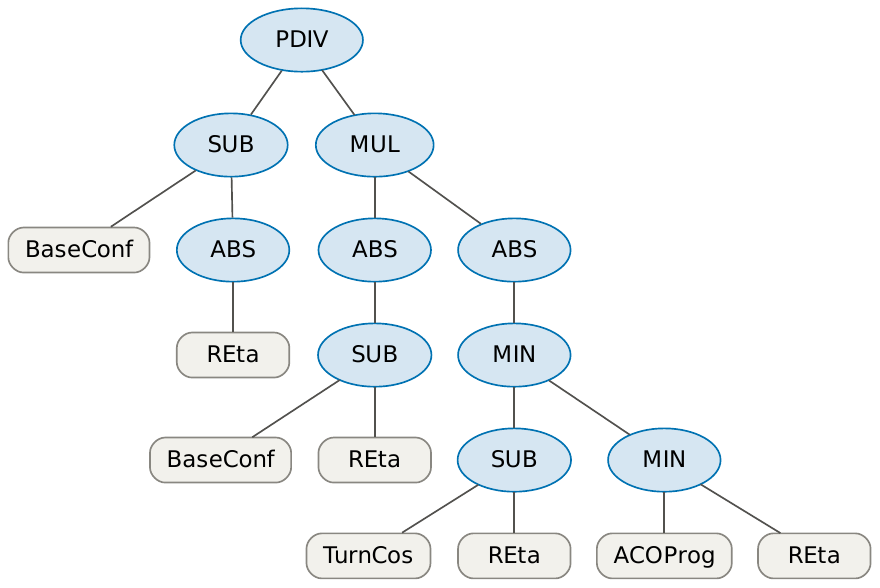}
\caption{Expression tree of the GP-ACO champion rule with the lowest validation gap (12.324\%) among the 10 independent runs on TSP500.}
\label{S-fig:best_tree}
\end{figure}

Fig.~\ref{S-fig:best_tree} shows the champion rule with the lowest validation gap among the 10 independent runs on TSP500. Ellipses are function nodes and rounded rectangles are terminals (defined in Table~\ref{S-tab:language}). The output $z_j$ of the tree modifies the selection weight of the candidate by Eq.~(\ref{eq:transition}) of the main paper.

\end{document}